\documentclass{article}

\usepackage[main, final]{neurips_2025}
\usepackage[utf8]{inputenc} % allow utf-8 input
\usepackage[T1]{fontenc}    % use 8-bit T1 fonts
\usepackage{hyperref}       % hyperlinks
\usepackage{url}            % simple URL typesetting
\usepackage{booktabs}       % professional-quality tables
\usepackage{amsfonts}       % blackboard math symbols
\usepackage{nicefrac}       % compact symbols for 1/2, etc.
\usepackage{microtype}      % microtypography
\usepackage{xcolor}         % colors

\usepackage{bm}
\usepackage{algorithm,algpseudocode}
\usepackage{amssymb}
\usepackage{amsmath}
\usepackage[inline]{enumitem}
\usepackage{nicematrix}
\usepackage{graphicx}
\usepackage{subcaption}
\usepackage{multirow}
\usepackage{tikz}
\usepackage{pgfplots}
\pgfplotsset{compat=1.18}

\algnewcommand\algorithmicforeach{\textbf{for each}}
\algdef{S}[FOR]{ForEach}[1]{\algorithmicforeach\ #1\ \algorithmicdo}

\makeatletter
\let\OldStatex\Statex
\renewcommand{\Statex}[1][3]{%
  \setlength\@tempdima{\algorithmicindent}%
  \OldStatex\hskip\dimexpr#1\@tempdima\relax}
\makeatother

\DeclareUnicodeCharacter{2212}{-}
\title{Graph Diffusion that can Insert and Delete}

\author{%
  Matteo~Ninniri\\
  Department of Computer Science\\
  University of Pisa\\
  56127 Pisa (Italy) \\
  \texttt{matteo.ninniri@phd.unipi.it} \\
  \And
  Marco~Podda\\
  Department of Computer Science\\
  University of Pisa\\
  56127 Pisa (Italy) \\
  \texttt{marco.podda@unipi.it} \\
  \AND
  Davide~Bacciu\\
  Department of Computer Science\\
  University of Pisa\\
  56127 Pisa (Italy) \\
  \texttt{davide.bacciu@unipi.it} \\
}

\begin{document}

\maketitle

\begin{abstract}
Generative models of graphs based on discrete Denoising Diffusion Probabilistic Models (DDPMs) offer a principled approach to molecular generation by systematically removing structural noise through iterative atom and bond adjustments. However, existing formulations are fundamentally limited by their inability to adapt the graph size (that is, the number of atoms) during the diffusion process, severely restricting their effectiveness in conditional generation scenarios such as property-driven molecular design, where the targeted property often correlates with the molecular size.
In this paper, we reformulate the noising and denoising processes to support monotonic insertion and deletion of nodes. The resulting model, which we call \textsc{GrIDDD}, dynamically grows or shrinks the chemical graph during generation. 
\textsc{GrIDDD} matches or exceeds the performance of existing graph Diffusion Models on molecular property targeting despite being trained on a more difficult problem. 
Furthermore, when applied to molecular optimization, \textsc{GrIDDD} exhibits competitive performance compared to specialized optimization models.
This work paves the way for size-adaptive molecular generation with graph diffusion.
\end{abstract}

\section{Introduction}
Generating molecules conditioned on predefined structures or properties is a central endeavor in computational chemistry, with applications ranging from \textit{de novo} drug design \citep{you2018gpcn} to materials discovery \citep{Zhao2023}. Depending on the conditioning information, we distinguish two key learning tasks: \textit{property targeting}, aimed at generating molecules endowed with prespecified properties; and \textit{property optimization}, where the goal is to edit a given molecule to improve a target property while retaining its core structure. Unlike continuous data, generating graphs must account for their discrete and combinatorial connectivity. For molecules in particular, chemical constraints invalidate most atom-bond and atom-atom combinations, making this problem difficult and, therefore, actively researched. Despite these challenges, deep graph generators \citep{faez2021survey} have achieved striking success in the above-mentioned tasks, due to their ability to approximate complex molecular distributions and their flexibility in incorporating conditioning information. 
 
Based on the way they decode a graph from a latent representation, deep generative models for molecules can be broadly categorized as autoregressive or one-shot \citep{zhu2022a}. Autoregressive models build the graph sequentially (atom-by-atom as in \citet{you2018}, or fragment-by-fragment as in \citet{jybj:1}), but suffer from order dependence and linearly-scaling sampling time. One-shot methods generate all nodes and edges in a single pass enabling parallel sampling, but struggle to learn high-order interactions. Denoising Diffusion Probabilistic Models (DDPMs) \citep{hja:1} for graphs bridge these approaches by progressively removing Gaussian or discrete noise to recover the molecular graph. Indeed, DDPMs preserve permutation invariance and sampling parallelism while iteratively refining long-range interactions (e.g., ring closures), which traditional one-shot decoders are forced to learn in a single transformation. Moreover, DDPMs can be easily adapted to conditional generation with classifier-based \citep{dhariwal2021diffusion} or classifier-free \citep{hs:1} guidance.

One drawback of Diffusion Models (not limited to graphs) is that the sample size remains fixed throughout the generative process. While for different modalities this is less of a concern (e.g., for images it is equivalent to fixing the resolution beforehand), it becomes relevant when generating molecules, and more in general when modeling combinatorial data.
In practical terms, this restriction implies that the model \textit{by design} cannot add or remove atoms during generation, leading to two major drawbacks: \textit{a)} in property targeting, it is not possible to make the molecular structure responsive to properties which correlate with the size, e.g., when generating molecules with a specific molecular weight; \textit{b)} in property optimization, the model cannot optimize a given property by adapting the generated structure. Existing methods circumvent this limitation by sampling the graph size before the generative process starts. For example, \citet{vkswcf:1} and \cite{hoogeboom22a} sample different graph sizes from the empirical distribution of the training dataset, while \citet{npb:1} use an auxiliary classifier to predict the graph size from the conditioning information. Though these workarounds are sufficient for property targeting, they become ineffective for property optimization, where size is strictly related to the structure to optimize. \citet{ketata2025lift} adapts to property optimization by sampling additional nodes in the reverse process, but does not incorporate this procedure into the training process.

\begin{figure}
  \centering
  \begin{subfigure}[b]{0.10\textwidth}
    \centering
    \includegraphics[width=\textwidth, trim={0cm 0cm 0cm 1.2cm},clip]{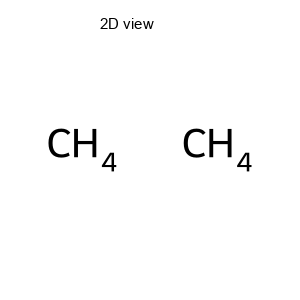}
  \end{subfigure}
  \begin{subfigure}[b]{0.10\textwidth}
    \centering
    \includegraphics[width=\textwidth, trim={0cm 0cm 0cm 1.2cm},clip]{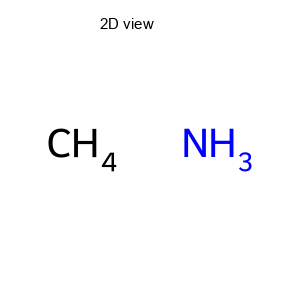}
  \end{subfigure}
  \begin{subfigure}[b]{0.10\textwidth}
    \centering
    \includegraphics[width=\textwidth, trim={0cm 0cm 0cm 1.2cm},clip]{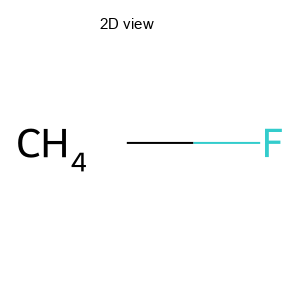}
  \end{subfigure}
  \begin{subfigure}[b]{0.10\textwidth}
    \centering
    \includegraphics[width=\textwidth, trim={0cm 0cm 0cm 1.2cm},clip]{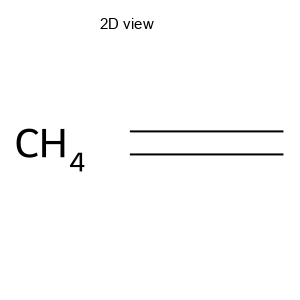}
  \end{subfigure}
  \begin{subfigure}[b]{0.10\textwidth}
    \centering
    \includegraphics[width=\textwidth, trim={0cm 0cm 0cm 1.2cm},clip]{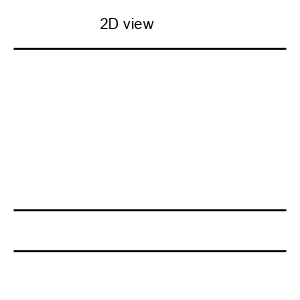}
  \end{subfigure}
  \begin{subfigure}[b]{0.10\textwidth}
    \centering
    \includegraphics[width=\textwidth, trim={0cm 0cm 0cm 1.2cm},clip]{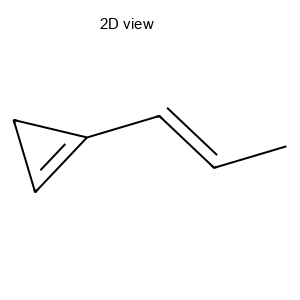}
  \end{subfigure}
  \begin{subfigure}[b]{0.10\textwidth}
    \centering
    \includegraphics[width=\textwidth, trim={0cm 0cm 0cm 1.2cm},clip]{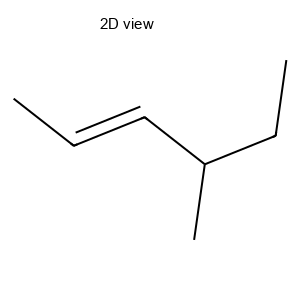}
  \end{subfigure}
  \begin{subfigure}[b]{0.10\textwidth}
    \centering
    \includegraphics[width=\textwidth, trim={0cm 0cm 0cm 1.2cm},clip]{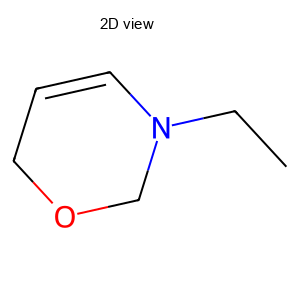}
  \end{subfigure}
  \begin{subfigure}[b]{0.10\textwidth}
    \centering
    \includegraphics[width=\textwidth, trim={0cm 0cm 0cm 1.2cm},clip]{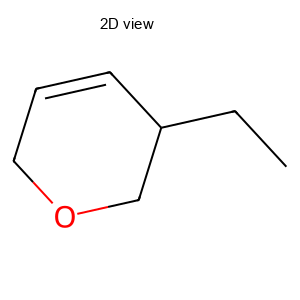}
  \end{subfigure}\\

  \begin{subfigure}[b]{0.10\textwidth}
    \centering
    \includegraphics[width=\textwidth, trim={0cm 0cm 0cm 1.2cm},clip]{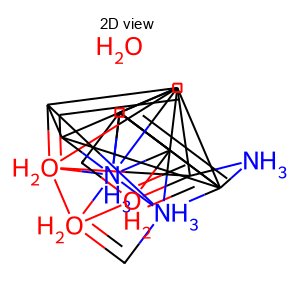}
  \end{subfigure}
  \begin{subfigure}[b]{0.10\textwidth}
    \centering
    \includegraphics[width=\textwidth, trim={0cm 0cm 0cm 1.2cm},clip]{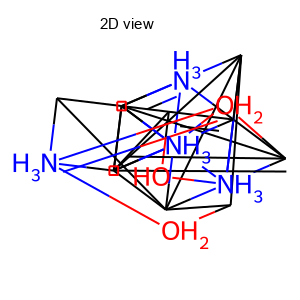}
  \end{subfigure}
  \begin{subfigure}[b]{0.10\textwidth}
    \centering
    \includegraphics[width=\textwidth, trim={0cm 0cm 0cm 1.2cm},clip]{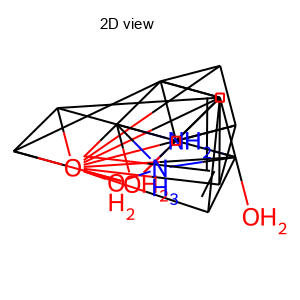}
  \end{subfigure}
  \begin{subfigure}[b]{0.10\textwidth}
    \centering
    \includegraphics[width=\textwidth, trim={0cm 0cm 0cm 1.2cm},clip]{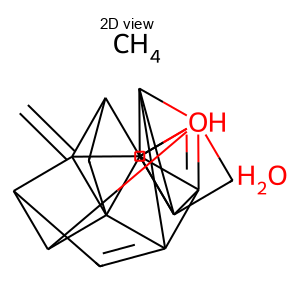}
  \end{subfigure}
  \begin{subfigure}[b]{0.10\textwidth}
    \centering
    \includegraphics[width=\textwidth, trim={0cm 0cm 0cm 1.3cm},clip]{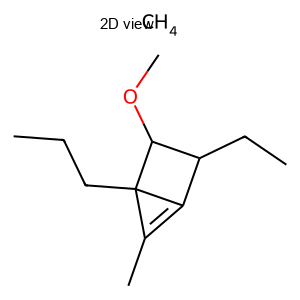}
  \end{subfigure}
  \begin{subfigure}[b]{0.10\textwidth}
    \centering
    \includegraphics[width=\textwidth, trim={0cm 0cm 0cm 1.2cm},clip]{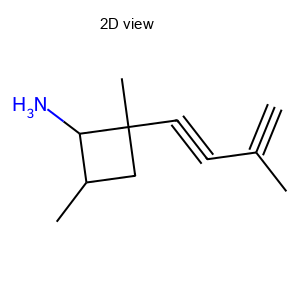}
  \end{subfigure}
  \begin{subfigure}[b]{0.10\textwidth}
    \centering
    \includegraphics[width=\textwidth, trim={0cm 0cm 0cm 1.2cm},clip]{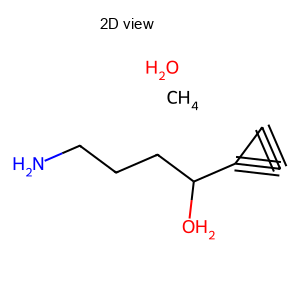}
  \end{subfigure}
  \begin{subfigure}[b]{0.10\textwidth}
    \centering
    \includegraphics[width=\textwidth, trim={0cm 0cm 0cm 1.2cm},clip]{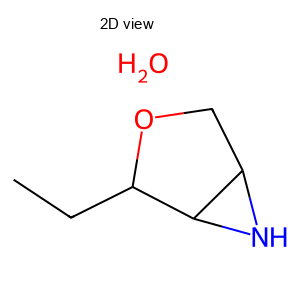}
  \end{subfigure}
  \begin{subfigure}[b]{0.10\textwidth}
    \centering
    \includegraphics[width=\textwidth, trim={0cm 0cm 0cm 1.2cm},clip]{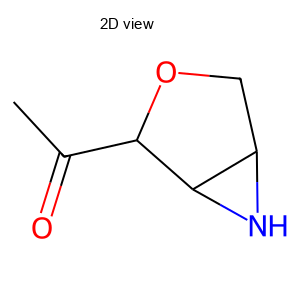}
  \end{subfigure}
  
  \caption{Two qualitative examples of the proposed model (\textsc{GrIDDD}) when generating molecules from the QM9 dataset. In the top row, from left to right, we show a subset of the denoising process to generate a molecule starting from a latent graph with two atoms (which are extremely rare in QM9). \textsc{GrIDDD} successfully inserts six more nodes to obtain a sample resembling the training set's distribution. In the bottom row, we start from a latent with 18 atoms instead (not present in QM9). \textsc{GrIDDD} manages to delete nine atoms and obtain a valid molecule. Notice that, unlike current DDPMs for graphs, the graph size is changed dynamically during denoising.}\label{fig:denoising}
\end{figure}

Motivated by these considerations, our primary contribution is a generalization of the standard discrete diffusion process on graphs. Our approach enables step-wise monotonic node insertions and removals during diffusion, and is specifically designed to change the graph size dynamically throughout the generative process, providing the necessary flexibility to incorporate conditional information more effectively. We implement this novel formulation as a model called \textsc{GrIDDD} (short for Graph Insert-Delete Discrete Diffusion). An intuitive example of how \textsc{GrIDDD} generates molecules is provided in Figure \ref{fig:denoising}. We test \textsc{GrIDDD} on property targeting in two widely used benchmarks (QM9 and ZINC-250k), where it consistently performs on par or better than the state of the art in terms of approximating the target property while keeping high chemical validity, despite having been trained on a more difficult problem. 
When applied to molecular optimization, \textsc{GrIDDD} convincingly outperforms other molecular optimizers, achieving a higher average improvement and optimization success rate.
To the best of our knowledge, this is the first work that allows size-adaptive molecular generation with graph diffusion. Our code is available at \href{https://github.com/mninniri/GrIDDD}{https://github.com/mninniri/GrIDDD}.

\section{Background and related works}
\paragraph{Notation}
We represent molecules with $n$ atoms as graphs $G = (\bm{X}, \bm{E})$, where $\bm{X} \in \mathbb{R}^{n \times a}$ is the matrix of one-hot encoded atom types (over $a$ possible atom types), and $\bm{{E}} \in \mathbb{R}^{n \times n \times b}$ is an adjacency (or edge) tensor that encodes both the bond connectivity and the one-hot encoded bond types (over $b$ possible bond types while considering the absence of a bond as an additional bond type). We denote as $\bm{x}_i \in \mathbb{R}^a$ the $i$-th row of $\bm{X}$ and as $\bm{e}_{ij} \in \mathbb{R}^b$ the slice of $\bm{E}$ along the first and second dimensions.  Oftentimes, we use the terms ``nodes'' and ``edges'' to refer to atoms and bonds, respectively. Conditioning vectors will be generally denoted as $\bm{y} \in \mathbb{R}^d$, $d$ being a positive integer. 

\subsection{Graph Generative Models for Property Targeting and Optimization}

Several models have recently been developed for property targeting, mostly based on graph diffusion since it is particularly suited for conditional generation. For example, DiGress \citep{vkswcf:1} uses a categorical diffusion process on nodes and edges, and a graph transformer denoiser to learn a generative distribution over molecular graphs. DiGress allows conditional sampling through classifier-based guidance, by training an auxiliary regressor to steer the generation towards the target properties. FreeGress \citep{npb:1} builds upon DiGress and proposes a classifier-free approach for conditional generation that injects the guide directly into the denoiser during training, greatly improving property-targeting accuracy.

Molecular property optimization is mainly achieved through  translation models and optimization models. VJTNNs \citep{jybj:1}, Seq2Seq models \citep{hysnjtce:1}, and HierG2G \citep{jbj:2} all work by translating between different types of molecular representations, ranging from junction trees to SMILES strings. While these models achieve good results, they require a dataset of pairs of similar molecules with specific properties, which are of limited availability. At the same time, they are not designed to generate new molecules from scratch. 
Optimization models are usually designed for molecular generation. GCPN \citep{you2018gpcn} and JT-VAE \citep{jbj:1} employ different techniques, ranging from reinforcement learning to junction trees, to generate data, and are easily extensible to perform property targeting and optimization.

\subsection{Discrete Denoising Diffusion Probabilistic Models}
DDPMs consist of an un-parameterized \textit{forward process} $q(\bm{x}^t|\bm{x}^{t-1})$ and a parameterized \textit{reverse process} $p_\theta(\bm{x}^{t-1}|\bm{x}^t)$. The forward process progressively corrupts the initial data point $\bm{x}^0$, transforming it into a noise-like sample $\bm{x}^T$. The reverse process, trained to denoise these samples, aims to reconstruct $\bm{x}^0$ by sequentially removing the added noise. Sampling from a trained DDPM involves drawing $\bm{\tilde{x}}^T$ from the noise distribution and applying the reverse process iteratively over $T$ steps to obtain a new sample $\bm{\tilde{x}}^0 \sim q(\bm{x})$. While in the original formulation of DDPMs the noise distribution is Gaussian, we focus on \textit{discrete} DDPMs where noise is injected through specially crafted \textit{transition matrices}  $\bm{Q}^t$, with $[\bm{Q}^t]_{ij} = p(x^t=j|x^{t-1}=i)$, encapsulating the probability of switching from  category $i$ to category $j$ at step $t$ of the forward process. Crucially, cumulative transitions from an arbitrary timestep $s$ to $t$ can be expressed as a matrix $\overline{\bm{Q}}_{t|s} = \prod_{i=s+1}^t \bm{Q}^i$. The reverse process $p_\theta(\bm{x}^{t-1}|\bm{x}^t)$ is computed by marginalization over the possible categories of the input $\bm{x}$. 

\subsection{Insert and delete operations for Diffusion Models}
The idea of insert and delete operations for discrete Diffusion Models originates from \citet{johnson2021beyond}, which, similarly to us, proposed to gradually insert or delete nodes both during the forward and reverse process. However, there are some major differences with our work. Firstly, our solution is designed for graphs, instead of textual data. Secondly, they resort to complex edit summaries to compute the various probability distributions involved, while we simplify the computation of the posteriors by generalizing the denoising process to account for nodes that have yet to be deleted or inserted through the computation.
Some continuous-time Diffusion Models \citep{jump_diffusion} can increase the entries in the data point during the reverse process, but can not remove them, making them unsuitable for property optimization. At the same time, they are better suited for continuous data with positional dependencies, rather than graphs.

\section{Graph Insert-Delete Discrete Diffusion}
We now introduce our main contribution. The backbone of \textsc{GrIDDD} is a reformulation of the standard discrete denoising diffusion for graphs to support the use of monotonic node insertions and deletions, which we describe in this section.
\paragraph{Preliminaries.} Our end goal is to obtain a denoised graph $\bm{G}^0 = (\bm{X}^0,\bm{E}^0)$ with $n^0$ nodes from a latent noisy graph with $\bm{G}^T = (\bm{X}^T,\bm{E}^T)$ with $n^T$ nodes. %The simplest way to 
To change size from $n^T$ to $n^0$, we \textit{monotonically} delete (if $n^T>n^0$) or insert (if $n^T < n^0$) $|\Delta^T|$ nodes, where $\Delta^T = n^T-n^0$, as the reverse process advances. This gives us full control over the graph size at any given moment. The timesteps when insertions and deletions are performed are sampled from the normalization of the absolute value of the logistic distribution's probability density function $\zeta^{'}(t)$:
\begin{equation} \label{eq:zeta_t}
    %\zeta^{'}(t) =-\frac{e^{\frac{t-D}{w}}}{w(e^{\frac{t-D}{w}}+1)^2}.
    \zeta^{'}(t) =|\frac{e^{\frac{t-D}{w}}}{w(e^{\frac{t-D}{w}}+1)^2}|.
\end{equation}
The function is scaled in a way such that  $\sum_{t=0}^T\zeta^{'}(t) = 1$, with $\zeta^{'}(0) = \zeta^{'}(T) = 0$. The parameter $D$ indicates the timestep in which the function is maximized, while $w$ is the scale parameter (from a visual standpoint, it controls the steepness of the curve). During training, the final size $n^T$ that a training sample with size $n^0$ will assume is sampled from a discrete distribution on the number of nodes $h_{n^0}(n)$. In our work, we defined it as follows, and then normalized it to sum to one:
\begin{equation}\label{eq:h_n0}
    h_{n^0}(n) = p_{\text{max}} + \frac{p_{\text{min}}-p_{\text{max}}}{n_{\text{max}}}|n -n^0|,
\end{equation}
where $n_{\text{max}}$ is the maximum size that a graph may assume (which can be set to be higher than the size of the largest graph in the dataset), while $p_{\text{min}}$ and $p_{\text{max}}$ are hyperparameters.
Intuitively, the highest probability $p_{\text{max}}$ occurs when $n = n^0$, and it linearly decreases to $p_{\text{min}}$ with a rate equal to $\frac{p_{\text{min}}-p_{\text{max}}}{n_{\text{max}}}$ as the absolute difference between $n^0$ and $n$ increases. 

\subsection{Forward process}
The discussion above implies that we have three different forward processes, depending on whether $\Delta^T > 0$,  $\Delta^T < 0$, or $\Delta^T = 0$. We start by describing the latter, which is the simplest and serves as a basis for the deletion and insertion cases. Building on previous works \citep{vkswcf:1,npb:1}, when $\Delta^T = 0$, the forward process is implemented as:
\begin{equation}
    q(\bm{G}^t\mid\bm{G}^{t-1}) = (\bm{X}^{t-1}\bm{Q}_{\bm{X}}^{t},\bm{E}^{t-1}\bm{Q}_{\bm{E}}^{t}).
\end{equation}
Focusing on the nodes $\bm{X}$ (edges are defined analogously), the associated transition matrix is: 
\begin{equation} \label{eq:old_qt}
    \bm{Q}_{\bm{X}}^t = \alpha^t\bm{A}_{\bm{X}} + (1-\alpha^t)\bm{B}_{\bm{X}},
\end{equation}
where $\bm{A}_{\bm{X}} = \bm{I}$, $\bm{B}_{\bm{X}} = \mathbf{1}_a\bm{m}_{\bm{X}}^{\textsf{T}}$, with $\bm{m}_{\bm{X}}$ being the vector of the marginal distribution of the atom types computed on the training set, and $\{\alpha_t\}_{i=1\dots T}$ being a set of \textit{noise schedulers}, chosen such that $\alpha_1 = 1$, $\alpha_T = 0$, and the intermediate elements gradually decrease from one to zero. Multi-step transitions from an arbitrary timestep $s$ can be computed as a matrix $\overline{\bm{Q}}_{\bm{X}}^{t|s}$ defined as:
\begin{equation} \label{eq:old_qts}
    \overline{\bm{Q}}_{\bm{X}}^{t|s} = \prod_{i=s+1}^t \bm{Q}_{\bm{X}}^i = \overline{\alpha}^{t|s}\bm{A}_{\bm{X}} + (1-\overline{\alpha}^{t|s})\bm{B}_{\bm{X}},
\end{equation} 
where $\overline{\alpha}^{t|s} = \frac{\overline{\alpha}^{t|0}}{\overline{\alpha}^{s|0}}$, with $\overline{\alpha}^{t|0}=\prod_{i=1}^t \alpha^i = \cos{(\frac{\pi}{2}\frac{(\frac{t}{T}+s)^\nu}{1+s})^2} $ (notice the hyperparameter $\nu$).

\paragraph{Monotonic deletions ($\Delta^T < 0$).}
To introduce deletions, we build on \citet{johnson2021beyond} and treat them as an additional atom type \texttt{DEL}. This is similar to absorbing diffusion \citep{DBLP:conf/icml/KongCSZPZ23}, in which the forward process consists of progressively switching all atom types to the deletion type; our strategy is more general, as we allow for nodes and edges to change category during the forward and reverse processes. 
During the reverse process, nodes need to be reinserted at step $t$ whenever they were deleted during the forward process at the same timestep (that is, if they transitioned to type \texttt{DEL} at forward step $t$). Consequently, such nodes are expected to revert to a proper (non-\texttt{DEL}) type at the next step $t-1$. However, since the \texttt{DEL} state is absorbing (meaning $p(\bm{x}^t \neq \texttt{DEL} \mid \bm{x}^{t-1} = \texttt{DEL}) = 0$), 
such transition is forbidden: once a node is deleted, it must remain deleted up to the end of the forward process, as doing otherwise would violate monotonicity. To enable node reinsertion, we introduce an auxiliary transient type \texttt{DEL}$^{*}$, designed such that $p(\bm{x}^t=\text{\texttt{DEL}}| \bm{x}^{t-1}=\text{\texttt{DEL}}^{*}) = 1$, i.e., the transition from \texttt{DEL}$^{*}$ to \texttt{DEL} during the forward process is deterministic, and $p(\bm{x}^{t-1} \in \{\text{\texttt{DEL}}, \text{\texttt{DEL}}^{*}\}| \bm{x}^{t}=\text{\texttt{DEL}}^{*}) =0$. This ensures that if a node is in state \texttt{DEL}$^{*}$ during the reverse process, it can only switch to a proper (non-\texttt{DEL}) type. This mechanism guarantees that nodes deleted at step $t$ can be reinserted at step  $t−1$, thereby preventing the absorbing \texttt{DEL} state from blocking node recovery. 
To ensure that the final graph size is consistent with the one fixed at the start of the forward process, preventing that too many nodes switch into the deletion state, we employ a hybrid scheme, in which $n_0 - |\Delta^T|$ nodes undergo the standard forward process (described by Eq. \ref{eq:old_qt}),  while $|\Delta^T|$ nodes use a different transition matrix\footnote{From here on, we drop the subscript $\bm{X}$ for brevity. Equations 6--9 are defined analogously for $\bm{E}$.} defined as:
\begin{equation}
    \bm{Q}^{*t} = \zeta(t)\left(\alpha^t\bm{A}^{*} + (1-\alpha^t)\bm{B}^{*}\right) + \left(1-\zeta(t)\right)\bm{C}^{*}.
\end{equation}
Above, $\bm{A}^{*}$ and $\bm{B}^{*}$ are $\bm{A}$ and $\bm{B}$ augmented to account for the deletion types, $\bm{C}^{*}$ describes the transition from a normal type to a deletion (see Figure \ref{fig:new_abcd}), and $\zeta(t)$ is the integral of $\zeta^{'}(t)$ described in Equation \ref{eq:zeta_t}. Since $\zeta^{'}(t)$ is a probability distribution, $\zeta(t)$ can be seen as its cumulative distribution function; therefore, at time $T$, $\bm{Q}^{*T} = \bm{C}^{*}$.
If $\zeta^{'}$ is chosen such that $\zeta(T-1) = 0$, we enforce exactly $|\Delta^T|$ nodes with  \texttt{DEL} type at timestep $T$. We refer the reader to Appendix \ref{app:zeta} for more details on the function $\zeta^{'}$ used in this study.
Computing the cumulative transition matrix $\overline{\bm{Q}}^{*t|s} = \prod_{i=s+1}^t \bm{Q}^{i*}$ requires an additional matrix $\bm{D}^{*}$ (also displayed in Figure \ref{fig:new_abcd}):
\begin{equation}\label{eq:Q_ts_star}
    \overline{\bm{Q}}^{*t|s} = \overline{\zeta}^{t|s}\left(\overline{\alpha}^{t|s}\bm{A}^{*} + \left(1-\overline{\alpha}^{t|s}\right)\bm{B}^{*}\right) + \overline{\zeta}^{t-1|s}\left(1-\zeta(t)\right)\bm{C}^{*} + \left(1-\overline{\zeta}^{t-1|s}\right)\bm{D}^{*},
\end{equation}
where $\overline{\zeta}^{t|s} = \prod_{i = s+1}^t \zeta(i)$. Intuitively, $\bm{D}^{*}$ represents a state where all nodes are immediately switched to state \texttt{DEL}. If a node is set to category \texttt{DEL} or \texttt{DEL}$^{*}$, so are the edges associated with it.

\begin{figure}
\noindent
\begin{minipage}{0.48\textwidth}
    \[
        \bm{A}^{*}=
		\begin{bNiceArray}{ccc:c:c}[margin,last-col,first-row]
            & & & \texttt{D} & \texttt{D}^{*} & \\
			\NotEmpty \Block{3-3}<\Huge>{\bm{A}}& & & 0 & 0 &\\
			  &       &   & \vdots & \vdots &\\
			  &       &   & 0 & 0 &\\
            \hdottedline
			0 & \cdots & 0 & 1 & 0 & \texttt{D}\\
            \hdottedline
			0 & \cdots & 0 & 1 & 0 & \texttt{D}^{*}\\
		\end{bNiceArray}
        \tag*{\hspace{1em}(1)}
    \]
\end{minipage}
\hfill
\begin{minipage}{0.48\textwidth}
    \[
        \bm{B}^{*}=
		\begin{bNiceArray}{ccc:c:c}[margin,last-col,first-row]
            & & & \texttt{D} & \texttt{D}^{*} & \\
			\NotEmpty \Block{3-3}<\Huge>{\bm{B}}& & & 0 & 0\\
			  &       &   & \vdots & \vdots\\
			  &       &   & 0 & 0\\
            \hdottedline
			0 & \cdots & 0 & 1 & 0 & \texttt{D}\\
            \hdottedline
			0 & \cdots & 0 & 1 & 0 & \texttt{D}^{*}\\
		\end{bNiceArray}
        \tag*{\hspace{1em}(2)}
    \]
\end{minipage}
\vspace{1em}
\noindent
\begin{minipage}{0.48\textwidth}
    \[
        \bm{C}^{*}=
		\begin{bNiceArray}{ccc:c:c}[margin,last-col,first-row]
            & & & \texttt{D} & \texttt{D}^{*} & \\
			0 & \cdots & 0 & 0 & 1\\
			\vdots & \ddots & \vdots & \vdots & \vdots\\
			0 & \cdots & 0 & 0 & 1\\
            \hdottedline
			0 & \cdots & 0 & 1 & 0 & \texttt{D}\\
            \hdottedline
			0 & \cdots & 0 & 1 & 0 & \texttt{D}^{*}\\
		\end{bNiceArray}
        \tag*{\hspace{1em}(3)}
    \]
\end{minipage}
\hfill
\begin{minipage}{0.48\textwidth}
    \[
        \bm{D}^{*}=
		\begin{bNiceArray}{ccc:c:c}[margin,last-col,first-row]
            & & & \texttt{D} & \texttt{D}^{*} & \\
			0 & \cdots & 0 &   1    & 0\\
			\vdots & \ddots & \vdots & \vdots & \vdots\\
			0 & \cdots & 0 &   1    & 0\\
            \hdottedline
			0 & \cdots & 0 & 1 & 0 & \texttt{D}\\
            \hdottedline
			0 & \cdots & 0 & 1 & 0 & \texttt{D}^{*}\\
		\end{bNiceArray}
        \tag*{\hspace{1em}(4)}
    \]
\end{minipage}
    
\caption{The matrices employed in the computation of $\bm{Q}^{*t}$ and $\overline{\bm{Q}}^{*t|s}$. For spacing reasons, the states \texttt{DEL} and \texttt{DEL}$^{*}$ have been shortened as \texttt{D} and \texttt{D}$^{*}$.}
\label{fig:new_abcd}
\end{figure}

\paragraph{Monotonic insertions ($\Delta^T > 0$).}
Node insertions are defined implicitly (i.e., without a specific \texttt{INS} type) since by design, a node is effectively inserted in the graph, or ``activated'', only at timestep $s+1$ if the model sampled $s$ as insertion time.
The critical operation after inserting nodes is to establish which category they belong to.
Our solution is to sample the inserted node's state from the marginal distribution $\bm{m_X}$ of the graph itself, which can be precomputed with no significant computational overhead (for more information, refer to Appendix \ref{app:labels}).
Once a node is inserted, so are the edges connecting it to the rest of the graph. Their initial state is computed analogously using $\bm{m}_{\bm{E}}$. 

\subsection{Reverse process}
The reverse process starts from a randomly initialized graph and gradually removes noise. However, since our main objective is to perform conditional generation, we also input the conditioning vector $\bm{y}$ using classifier-free guidance \citep{hs:1}. We factorize the reverse process as the product of the individual reverse probabilities of the nodes and edges (assuming mutual independence, which we discuss in Appendix \ref{app:node_edge_indip}):
\begin{equation}
    p_\theta(\bm{G}^{t-1}|\bm{G}^t, \bm{y}) = \prod_{i \in 0,\dots, n^t-1}p_\theta(\bm{x}^{t-1}_i|\bm{x}^t_i, \bm{y}) \prod_{i,j \in 0,\dots, n^t-1} p_\theta(\bm{e}^{t-1}_{ij}|\bm{e}^t_{ij}, \bm{y}).
\end{equation}

In previous works \citep{npb:1}, the posterior for the nodes was implemented with the aid of a neural network $p_\theta(\bm{x}^0 = \bm{x}| \bm{x}^t,\bm{y})$ tasked to predict the true atom and bond types given their noisy categories and the guide. However, this no longer works in our setting, as we cannot predict nodes that did not exist when $t=0$ but were inserted subsequently during the forward process. To address this issue, we let the main model predict, alongside $\bm{G}^0$, the activation time $\hat{s}$ of each node and introduce the following generalized posterior:
\begin{align} \label{eq:new_posterior}
    p_\theta(\bm{x}^{t-1}|\bm{x}^{t},\bm{y}) 
&= \sum_{\bm{x} \in \chi} \frac{q(\bm{x}^{t}|\bm{x}^{t-1}) q(\bm{x}^{t-1}|\bm{x}^{\hat{s}} = \bm{x})}{q(\bm{x}^{t}|\bm{x}^{\hat{s}} = \bm{x})}\,p_\theta(\bm{x}^{\hat{s}} = \bm{x}| \bm{x}^t,\bm{y}).
\end{align}

The activation time of the nodes belonging to the original graph is assumed to be zero. 
Notice that when $q(\bm{x}^{t}|\bm{x}^{0} = \bm{x}) = 0$, then $p_\theta(\bm{x}^{t-1}|\bm{x}^t,\bm{y})$ is set to zero as well. The edge posterior is computed analogously. Below, we describe how to extend the base setting to support insertions and deletions.

\paragraph{Deletions.}
During the reverse process, nodes deleted during the forward process need to be re-inserted.
We do so with an auxiliary neural network $g_\phi(\bm{G}^t)$ that predicts, given the latent noisy graph $\bm{G}^t$, the number of \texttt{DEL}$^{*}$s to add before passing the model to the main neural network that predicts the original graph $\bm{G}^0$ given $\bm{G}^t$. The auxiliary network is trained alongside the main model by masking out the nodes set to \texttt{DEL}$^{*}$ in $\bm{G}^t$ after computing $p_\theta(\bm{G}^0|\bm{G}^t,\bm{y})$, which reduces the training time significantly. Notice that $g_\phi$ is trained only when $\zeta^{'}(t) > 0$, since it is not possible to insert nodes otherwise. 

\paragraph{Insertions.}
Conversely to the deletion case, nodes inserted during the forward process need to be deleted. To do this, we let the main model predict, alongside $\bm{G}^0$, the activation time $\hat{s}$ of each node. Then, at timestep $t$, we remove from the graph the nodes predicted to have been inserted at step $t$. The activation time of the nodes belonging to the original graph is assumed to be zero. 

\subsection{Training and sampling} 
\paragraph{Training.} During the learning phase, our goal is to train the model by corrupting each dataset sample $\bm{G}^0 = (\bm{X}^0,\bm{E}^0)$ into $\bm{G}^{*t} =  (\bm{X}^{*t}, \bm{E}^{*t})$, with $t \leq T$, sampled from a uniform distribution, using the forward process. Having been subject to insertions or deletions, $\bm{X}^{*t}$ might contain only a subset of the nodes in $\bm{X}^0$ (if we deleted nodes) or include a new set of nodes (the ones that have been inserted up to step $t$). We train the main model to predict the values that the nodes $\bm{X}^{*t}$ and edges $ \bm{E}^{*t}$ had at their respective activation times $\bm{S}^{*}$ (which may be equal to zero if the element belongs to the original graph) and denote these prediction as $\widehat{\bm{X}}^{*0}$ and $ \widehat{\bm{E}}^{*0}$. At the same time, we predict the activation times themselves, which we denote as $\widehat{\bm{S}}^{*}$. Finally, we train the auxiliary model $g_\phi(\bm{G}^{*t})$ to predict the correct amount $n_{\texttt{DEL}^{*}}$ of \texttt{DEL}$^{*}$s that were present in $\bm{G}^{*t}$, which we denote as $\widehat{n}_{\texttt{DEL}^{*}}$. The final loss is a sum of Cross-Entropy (CE) terms of the targets $\bm{X}^{*t}, \bm{E}^{*t}, \bm{S}^ {*}$, and $n_{\texttt{DEL}^{*}}$ against their respective predictions:
\begin{equation}
    \lambda_X \mathrm{CE}(\widehat{\bm{X}}^{*0}, \bm{X}^{*0}) + \lambda_E \mathrm{CE}(\widehat{\bm{E}}^{*0}, \bm{E}^{*0}) + \lambda_S \mathrm{CE}(\widehat{\bm{S}}^{*}, \bm{S}^{*}) + \lambda_{\texttt{DEL}}\mathrm{CE}(\widehat{n}_{\texttt{DEL}^{*}}, {n}_{\texttt{DEL}^{*}}),
\end{equation}
where $\lambda_X, \lambda_E, \lambda_E, $ and $\lambda_{\texttt{DEL}}$ are hyperparameters that control the weight of each term. If we are training a model to condition on a specific set of properties $\bm{y}$, it is also fed at training time an auxiliary term $\bm{c}$ which, with a probability equal to $1-\rho$, $\rho$ being an hyperparameter, will be equal to $\bm{y}$, while in the remaining cases it will assume the value of a parametrized placeholder $\overline{\bm{y}}$, which is randomly initialized. This technique is called \textit{conditional dropout} and is shown to produce better results in conditional generation \citep{hs:1}. Algorithm \ref{alg:new_training} in Appendix \ref{app:algos} describes the new training process.

\paragraph{Sampling.}
During sampling, we start from a random latent $\bm{G}^T$ sampled from the marginal distributions on the nodes and edges $\bm{m_X}$ and $\bm{m_E}$. The size of a sample is also chosen from the marginal distribution of the nodes in the training set. At each step, we first insert as many \texttt{DEL}$^{*}$ nodes as predicted by the auxiliary neural network. Then, we feed the current latent $\bm{G}^t$ to the main model to predict $\bm{G}^{*0}$. That is, the values that each node and edge had at their respective activation timesteps, as well as the activation timesteps themselves. Afterwards, we delete the nodes that are predicted to have been inserted at timestep $t$. Finally, we use these information to compute Equation \ref{eq:new_posterior}, and sample from it the node and edge values at step $t-1$ to obtain $\bm{G}^{t-1}$. We then repeat the operations using it as the new $\bm{G}^t$, and we keep doing so for a total of $T$ steps until we obtain the final graph $\bm{G}^0$. In the case in which we are performing conditional generation on a property $\bm{y}$, the term $p_\theta(\bm{x}^{\hat{s}}|\bm{x}^t)$ is re-elaborated as follows:
\begin{equation}
    p_\theta(\bm{x}^{\hat{s}} = \bm{x}| \bm{x}^t,\bm{y}) = p_\theta(\bm{x}^{\hat{s}} = \bm{x}| \bm{x}^t, \overline{\bm{y}}) + \lambda (p_\theta(\bm{x}^{\hat{s}} = \bm{x}| \bm{x}^t,\bm{y}) - p_\theta(\bm{x}^{\hat{s}} = \bm{x}| \bm{x}^t,\overline{\bm{y}})),
\end{equation}
where $\lambda$ is a hyperparameter controlling the influence of the conditioning vector $\bm{y}$. Algorithm \ref{alg:new_denoising} in Appendix \ref{app:algos} describes the sampling process in detail.

\section{Experiments}
Here, we detail the experiments to evaluate \textsc{GrIDDD} on different property targeting and optimization tasks. Our code is based on MiDi \citep{midi} which, in turn, is based on DiGress. In all our experiments, we have $T = 500$ diffusion timesteps. The function $\zeta^{'}(t)$ is parameterized with $w=0.05$ and $D=\frac{T}{2}=250$. Similarly to MiDi, we use $\nu = 1$ for the node matrices' noise schedulers and $\nu = 1.5$ for the edge matrices. $\lambda_X$ and $\lambda_E$ are set, respectively, to 1 and 2. We set the hyperparameters $p_{\text{min}}$ and $p_{\text{max}}$ in Equation \ref{eq:h_n0} respectively as 0.2 and 1.
 
\paragraph{Datasets.}
Following previous works \citep{npb:1,vkswcf:1}, we used QM9 \citep{Ramakrishnan2014}, a dataset of 133k molecules made by up to 9 non-hydrogen atoms, and ZINC-250k, a collection of 250k drug-like molecules selected from the ZINC dataset \citep{is:1}. On ZINC-250k, the molecules were first preprocessed by removing stereochemistry information and infrequent non-neutrally-charged atoms, leaving only N+ and O-. These atoms have been treated as standalone atom types. Dataset splits are described in Appendix \ref{app:splits}.

\subsection{Property Targeting} \label{sec:targeting}
We adhere to the setup of \citet{npb:1}, where they extract 100 property vectors from the test set and generate, for each one of them, 10 molecules. Then, they compute the Mean Absolute Error (MAE) between the target properties $\bm{y}$ and the estimated properties of the generated molecules $\widehat{\bm{y}}$:
\begin{equation}
    \mathrm{MAE}(\bm{y}, \widehat{\bm{y}}) = \frac{1}{1000}\sum_{i=1}^{100}\sum_{j=1}^{10}|\bm{y}_i - \widehat{\bm{y}}_{i,j}|.
\end{equation}
On QM9, we targeted the Dipole Moment $\mu$ and the Highest Occupied Molecular Orbital (HOMO) properties, while on ZINC-250k we targeted the Log-Partition coefficient (LogP), the Quantitative Estimation of Drug-likeness (QED), and the molecular weight (MW). Since MW strongly correlates with the number of nodes of the molecular graph, it provides a natural and intuitive testbed to assess the ability to flexibly adapt the graph size, which is precisely what \textsc{GrIDDD} was designed for. Our baseline for comparison is discussed in Appendix \ref{app:baseline}.

\subsection{Property Optimization} \label{sec:optimization}
We use the test suite by \citet{jbj:2}, which consists of three experiments on the ZINC-250k dataset. The objective is to generate molecules that improve on a certain property while meeting a similarity constraint. 
To optimize a molecule, we corrupt it for 100 steps before starting the denoising process, setting the property vector $\bm{y}$ as the target value that the optimized molecule should have. 
In the \textbf{LogP} experiment, we optimize the 800 molecules with the lowest LogP among those in the test set. For each of them, we sample 20 candidates starting from different latents and only keep the one with the largest improvement among the ones with a fingerprint Tanimoto similarity with the starting molecule equal to or above a threshold $\delta$. We report the average of these improvements across the 800 molecules, with $\delta \in \{0.4, 0.6\}$.
In the \textbf{QED} experiment, we optimize 800 molecules from the test set with QED in the range $[0.7, 0.8]$. For each molecule, we sample 20 candidates from different latents. We consider the optimization successful if at least one among the candidates has a QED in the range $[0.9,1.0]$ and a fingerprint Tanimoto similarity with the starting molecule $\geq 0.4$. We report the success rate across the 800 molecules.
In the \textbf{DRD2} experiment, we optimize for the biological activity against the dopamine type 2 receptor, selecting 800 molecules from the test set with DRD2 activity score $\leq 0.05$. For each molecule, we sample 20 candidates from different latents. We consider the optimization successful if at least one among the candidates has a DRD2 score $\geq 0.5$ and a fingerprint Tanimoto similarity with the starting molecule $\geq 0.4$. We report the success rate across the 800 molecules. 
Our baseline for comparison is discussed in Appendix \ref{app:baseline}.

\subsection{Out-of-distribution sampling} \label{sec:ood}
To assess whether \textsc{GrIDDD} is able to generate valid molecules outside the training distribution, we train an instance of DiGress and one of \textsc{GrIDDD} to generate samples unconditionally on QM9. However, during training, we force \textsc{GrIDDD} to \textit{insert} up to 14 nodes in the molecular graph, despite the fact that QM9 features molecules with up to 9 atoms. Then, during generation, we force both models to generate molecules with up to 15 atoms (that is, above \textsc{GrIDDD}'s training capacity).

\paragraph{Experimental setup.}
The denoising network shares a similar architecture to FreeGress, and is described in Appendix \ref{sec:architecture}. For QM9, we use the same hyperparameters employed by the authors. The auxiliary neural network used to predict the number of \texttt{DEL}$^{*}$s also uses the exact same setup, with the exception that it only uses one layer. For ZINC-250k, we also use the same hyperparameters as FreeGress except for the number of layers, which we set to 10 instead of the original 12 to use approximately the same number of parameters. The size of the linear layer used to predict the activation time is 256. In QM9, we set the guidance scale to $\lambda = 3$, while in ZINC-250k we set $\lambda = 2$. These values have been chosen since they were the ones offering the best results for FreeGress. All experiments have been performed on an nVidia A100 GPU with 80 GBs of VRAM (two on ZINC-250k).

\section{Results}
Here, we detail the experimental results obtained by \textsc{GrIDDD}. A visual collection of samples obtained from the model is available in Appendix \ref{app:samples}
\subsection{Property targeting}
\paragraph{QM9.} Results are displayed in Table \ref{tab:results_qm9}. As can be seen, \textsc{GrIDDD} stably achieves state-of-the-art performance. When conditioning on $\mu$, it improves MAE by 14\% at the expense of a slightly lower validity. Conversely, when conditioning on HOMO, the model achieves a better validity at the expense of a slightly higher MAE. For ablation purposes, we also include an unconditional model in the comparison, which is unable to optimize the properties to a satisfactory degree. A discussion of the failure cases in QM9 is presented in Appendix \ref{app:failure}.

%\scriptsize
\begin{table}[ht]
    \centering
    \footnotesize
    \caption{Results of property targeting on QM9.}
    \label{tab:results_qm9}
    \begin{tabular}{lcccc}
        \toprule
        \multirow{2}{*}{\textbf{Method}}
        &\multicolumn{2}{@{}c@{}}{\textbf{$\mu$}}  &\multicolumn{2}{@{}c@{}}{\textbf{HOMO}} \\
        \cmidrule(lr){2-3} \cmidrule(lr){4-5}
        & \textbf{MAE $\downarrow$} & \textbf{Val. $\uparrow$} & \textbf{MAE $\downarrow$} & \textbf{Val. $\uparrow$}\\
        \midrule
        Unconditional & 1.68 $\pm$ 0.15 & 91.5 $\%$ & 0.95 $\pm$ 0.10 & 91.5 $\%$\\
        DiGress 
        & 0.80 $\pm$ 0.07 & 82.5 $\%$ & 0.61 $\pm$ 0.07 & 91.2 $\%$\\
        FreeGress
        & 0.74 $\pm$ 0.08 & \textbf{83.7 $\%$} & \textbf{0.32 $\pm$ 0.04} & 90.1 $\%$ \\
        \midrule
        \textsc{GrIDDD}
        & \textbf{0.66 $\pm$ 0.07} & 79.0 $\%$ & 0.37 $\pm$ 0.07 & \textbf{94 $\%$} \\
        \bottomrule
    \end{tabular}
\end{table}

\paragraph{ZINC-250k.} Results are shown in Table \ref{tab:results_zinc_single}. On ZINC-250k, \textsc{GrIDDD} once again achieves comparable or better MAE, while improving validity rates in two cases out of three. Of particular interest is the MW experiment, as this property strongly correlates with the graph size. Note that FreeGress' results have been obtained using an auxiliary neural network to predict the optimal number of atoms the molecule should have given the target MW, information which is used to set the graph's size before starting the denoising process.
\textsc{GrIDDD} is capable of halving FreeGress’ MAE without any prior knowledge of the graph size distribution. 
In summary, \textsc{GrIDDD} shows an improved capacity in generating valid molecules without sacrificing the MAE. What is truly relevant in these results is the fact that \textsc{GrIDDD} is capable of obtaining them despite being trained on a more difficult task, showing that our solution has a smoother tradeoff between validity and accuracy.

\begin{table}[htb!]
\centering
\footnotesize
\caption{Results of property targeting on the ZINC-250k dataset.}
\label{tab:results_zinc_single}
\begin{tabular}{lcccccc}

\toprule
\multirow{2}{*}{\textbf{Method}}
&\multicolumn{2}{@{}c@{}}{\textbf{LogP}}  &\multicolumn{2}{@{}c@{}}{\textbf{QED}} &\multicolumn{2}{@{}c@{}}{\textbf{MW}}\\
\cmidrule(lr){2-3} \cmidrule(lr){4-5} \cmidrule(lr){6-7}

& \textbf{MAE $\downarrow$} & \textbf{Val. $\uparrow$} & \textbf{MAE $\downarrow$} & \textbf{Val. $\uparrow$} & \textbf{MAE $\downarrow$} & \textbf{Val. $\uparrow$} \\
\midrule

Unconditional & 1.52 $\pm$ 0.12 & 86.1 $\%$ & 0.15 $\pm$ 0.01 & 86.1 $\%$ & 74.16 $\pm$ 6.71 & 86.1 $\%$\\

DiGress          &0.74 $\pm$ 0.08 & 74.6 $\%$ & 0.15 $\pm$ 0.01 & 85.1 $\%$ & 20.92 $\pm$ 2.90 & 40.4 $\%$\\
FreeGress       &\textbf{0.17 $\pm$ 0.01} & 84.9 $\%$ & 0.04 $\pm$ 0.01 & 84.9 $\%$ &  8.96 $\pm$ 1.93 & 79.7 $\%$\\
\midrule
\textsc{GrIDDD} &0.19 $\pm$ 0.02 & \textbf{87.9 $\%$} & \textbf{0.04 $\pm$ 0.00} & \textbf{87.2 $\%$} & \textbf{4.89 $\pm$ 0.57} & \textbf{84.2$\%$}\\
\bottomrule
\end{tabular}
\end{table}

\subsection{Out-of-distribution sampling}
Results are displayed in 
%Figure \ref{fig:ood}. 
Figure \ref{fig:ood_validity_qm9}.
While both models perform similarly with up to 12 atoms, the validity rates of DiGress steeply decrease on larger graph sizes. Notably, \textsc{GrIDDD} still generates 35$\%$ of valid molecules with 15 nodes, despite being trained on latents with up to 14 nodes. This shows how our model is fairly usable in out-of-distribution sampling, as it is able to learn graph sizes that do not appear in the training dataset.
\begin{figure}[h]
    \centering
    \begin{tikzpicture}
        \begin{axis}[
            width=0.9\linewidth,
            height=4cm,
            xlabel={Molecule size (atoms)},
            ylabel={Validity (\%)},
            ymin=0, ymax=110,
            xtick={10,11,12,13,14,15},
            grid=both,
            grid style={dashed,gray!20},
            legend style={
                at={(0.75,0.7)},
                anchor=south,
                legend columns=-1,
                /tikz/every even column/.append style={column sep=0.4cm}
            },
            every axis plot/.append style={
                thick,
                mark=*,
                mark size=2.2pt,
            },
        ]
        % DiGress (pastel blue)
        \addplot+[color=blue!60!white, mark options={fill=blue!60!white}] coordinates {
            (10,96.9)
            (11,85.8)
            (12,57.3)
            (13,30.2)
            (14,11.5)
            (15,5.9)
        };
        % GrIDDD (pastel red)
        \addplot+[color=red!60!white, mark options={fill=red!60!white}] coordinates {
            (10,90.0)
            (11,72.8)
            (12,54.3)
            (13,41.7)
            (14,38.6)
            (15,35.6)
        };
        \legend{\textsc{DiGress}, \textsc{GrIDDD}}
        \end{axis}
    \end{tikzpicture}
    \caption{Validity results of out-of-distribution sampling on QM9.}
    \label{fig:ood_validity_qm9}
\end{figure}

\subsection{Property optimization}
Results are summarized in Table \ref{tab:optimization}. 
Regarding the LogP property optimization, \textsc{GrIDDD} outperforms all competitors in both settings in terms of improvement. In particular, when the similarity threshold is 0.6, it achieves an average improvement almost twice as high as the second best. In the other two tasks, \textsc{GrIDDD} scores remarkably better, especially in the QED task where it improves 5 times the success rates of GCPN. An additional comparison with the method by \cite{ketata2025lift} (under a slightly different experimental setup) is provided in Appendix \ref{app:syco}, showing that \textsc{GrIDDD} performs comparably or better than methods that use 3D information to optimize molecules.
\begin{table}[ht]
\caption{Benchmark on graph property optimization. %{\color{red}The last row employs the setup used by Ketata et. al. \cite{ketata2025lift}, which is described in the Appendix, Section \ref{sec:ketata_setup}}.
}
\centering
\footnotesize
\renewcommand{\arraystretch}{1.2}
\begin{tabular}{@{}lcccccccc@{}}
\toprule
\multirow{2}{*}{\textbf{Method}} & 
\multicolumn{2}{c}{\textbf{LogP (sim $\geq$ 0.4)}} & 
\multicolumn{2}{c}{\textbf{LogP (sim $\geq$ 0.6)}} & 
\multicolumn{2}{c}{\textbf{QED (sim $\geq$ 0.4)}} & 
\multicolumn{2}{c}{\textbf{DRD2 (sim $\geq$ 0.4)}} \\

\cmidrule{2-9}
& \textbf{Improv. $\uparrow$} & \textbf{Div.$\uparrow$} & \textbf{Improv. $\uparrow$} & \textbf{Div.$\uparrow$} & \textbf{Succ.$\uparrow$} & \textbf{Div.$\uparrow$} & \textbf{Succ.$\uparrow$} & \textbf{Div.$\uparrow$} \\
%\midrule
%\multirow{4}{*}{\rotatebox[origin=c]{90}{\textsc{Transl.}}}
%& \textsc{Seq2Seq} &         3.37 $\pm$ 1.75 &          0.471  &         2.33 $\pm$ 1.17  & 0.331 & 58.5 $\%$ & 0.331 & 75.9 $\%$ & 0.176 \\
%& \textsc{JTNN   } &         3.55 $\pm$ 1.67 &          0.480  &         2.33 $\pm$ 1.24  & 0.333 & 59.9 $\%$ & 0.373 & 77.8 $\%$ & 0.156 \\
%& \textsc{AtomG2G} & \textbf{3.98 $\pm$ 1.54} & \textbf{0.563} &         2.41 $\pm$ 1.19  & 0.379 & 73.6 $\%$ & 0.421 & 75.8 $\%$ & 0.128 \\
%& \textsc{HierG2G} & \textbf{3.98 $\pm$ 1.46} &         0.381  & \textbf{2.49 $\pm$ 1.09} & \textbf{0.381} & 76.9 $\%$ & \textbf{0.477} & \textbf{85.9 $\%$} & 0.192 \\
\midrule
%\multirow{4}{*}{\rotatebox[origin=c]{90}{\textsc{Baseline}}}
\textsc{JT-VAE  }    & 1.03 $\pm$ 1.39 & -      & 0.28 $\pm$ 0.79 & - & 8.8 $\%$  & -     & 3.4 $\%$  & -     \\
\textsc{CG-VAE  }    & 0.61 $\pm$ 1.09 & -      & 0.25 $\pm$ 0.74 & - & 4.8 $\%$  & -     & -      & -     \\
\textsc{GCPN    }    & 2.49 $\pm$ 1.30 & -      & 0.79 $\pm$ 0.63 & - & 9.4 $\%$  & 0.216 & 4.4 $\%$  & \textbf{0.152} \\
%& \textsc{EDM-SyCo}    & 3.11 $\pm$ 1.27 & \textbf{0.555}  & 1.51 $\pm$ 1.10 & \textbf{0.360} & 46.4 $\%$ & 0.163 & \textbf{27.3 $\%$} & 0.083 \\

\midrule
%\multirow{2}{*}{\rotatebox[origin=c]{90}{\textsc{Ours}}}
\textsc{GrIDDD} & \textbf{2.70} $\pm$ 0.94  & 0.482 & \textbf{1.33} $\pm$ 0.61 & 0.280 & \textbf{45.1} $\%$ &   \textbf{0.283} & \textbf{5.0} $\%$  & 0.121 \\
%& \textsc{GrIDDD} (400 rounds) &  \textbf{3.27 $\pm$ 0.91} & 0.511 &  \textbf{1.59 $\pm$ 0.54} & \textbf{0.359} & \textbf{64.1 $\%$}  & \textbf{0.269} & 19.7 $\%$ & 0.058 \\
\bottomrule
\end{tabular}
\label{tab:optimization}
\end{table}

\subsection{Computational considerations}
The insertions and deletions performed by \textsc{GrIDDD} inevitably introduce computational overhead during training, leading to slightly increased training durations compared to methods that do not employ these operations. In particular, our analysis indicates that it is 30\% slower than FreeGress (see Appendix \ref{app:runtime} for an extended discussion). However, its flexibility in adapting the number of denoising steps to the task makes it generally faster to sample from (as detailed in Appendix \ref{app:sampling-speed}).

\paragraph{Ablation.} 
%{\color{red}\section{Ablation studies}}
To test the effective influence of insert and delete operations over the optimization process, we performed the same experiments with \textsc{GrIDDD} but disabling insertions and deletions. The results show that while the success rate remains relatively unchanged in the LogP and DRD2 experiments, it significantly drops to 33.8$\%$ when optimizing QED, likely because the QED score is a function of the molecular weight (and thus, it correlates with the number of atoms). We conclude that, similarly to targeting MW, controlling the graph size substantially increases success rates.

A naive form of insertion and deletion operations can be achieved implicitly on any graph DDPM by working with a large graph size (even beyond the largest in the training data), treating excess nodes/edges as padding nodes marked by a special \texttt{PAD} class which allows their removal at the end of the denoising process.  
Clearly, this method is slower to train and sample from, as it always works with maximally sized graphs. To check that this approach is also less accurate, we compared two DiGress variants that implement node and edge padding against \textsc{GrIDDD}: the first only assigns the \texttt{PAD} category to padding nodes, while the second also assigns it to their associated edges. All models were trained on the QM9 dataset for 250 epochs, after which we sampled 100 molecules each without conditioning. The results are displayed in Table \ref{tab:pad_ablation}. As it is possible to see, while the baseline generates $3\%$ more valid molecules than \textsc{GrIDDD},  they severely underperform in every other metric.
\begin{table}[!htbp]
    \centering
    \footnotesize
    %\begin{tabular}{l|cccccccc}
    \caption{Comparison of GrIDDD with DiGress-based padding variants that perform implicit insertions and deletions. Val: validity, Avg NC: average number of connected components, Max NC: maximum number of connected components, NSC: number of graphs sampled with a single connected component, XCE (ECE): validation cross-entropy over nodes (edges).} % The metrics shown are: validity, average number of connected components (NC), maximum NC among the sampled graphs, number of graphs sampled with one connected component, and validation Cross-Entropy (nodes and edges).}
    \begin{tabular}{lcccccc}
        \toprule 

         Model & \textbf{Val $\uparrow$} & \textbf{Avg NC $\downarrow$} & \textbf{Max NC $\downarrow$} & \textbf{NSC $\uparrow$} & \textbf{XCE $\downarrow$} & \textbf{ECE $\downarrow$} \\ 
         \midrule 
         GrIDDD    &   0.97   &  $\bm{1}$ & $\bm{1}$ & $\bm{100}$ & $\bm{0.44}$ & $\bm{0.32}$ \\ 
         Nodes \texttt{PAD} & $\bm{1}$ & 1.56      & 8 & 69 & 0.47 & 0.37 \\ 
         All \texttt{PAD}   & $\bm{1}$ & 1.2       & 4 & 84 & 0.48 & 0.38 \\
        \bottomrule
        \end{tabular}
    \label{tab:pad_ablation}

\end{table}

In Appendix \ref{app:sensitivity-analysis}, we conduct a sensitivity analysis on some critical hyperparameters of \textsc{GrIDDD}, showing that results are robust to their choices.

\section{Conclusions}
We introduced \textsc{GrIDDD}, a discrete DDPM which supports the insertion and deletion of atoms during the denoising process. This addresses one of the major limitations of graph DDPMs, which cannot change the graph size throughout the generative process. 
In future studies, we aim to study a simplified approach that does not need an auxiliary network to predict when to insert a node during the denoising process, and whether the distribution of the inserted nodes can be learned rather than employing predefined heuristics. 
Lastly, since the methodology we propose is quite general, we foresee its applications to domains such as, e.g., vector floor-plan generation \citep{Shabani2023}.

\paragraph{Limitations.}
During generation, we observed that the two neural networks composing \textsc{GrIDDD} may occasionally conflict, simultaneously performing an insertion and a deletion at the same timestep, which is theoretically illegal in our framework. \textsc{GrIDDD} also tends to produce more split molecules compared to FreeGress and DiGress. This is likely because nodes inserted late in the denoising process are challenging to connect before denoising concludes. Incorporating additional input features, such as the current number of connected components, could potentially mitigate this issue.

\section*{Acknowledgments} This work has been partially supported by EU-EIC EMERGE (Grant No. 101070918) and  PNRR, PE00000013, ``FAIR - Future Artificial Intelligence Research'', Spoke 1, funded by European Commission under NextGeneration EU programme (CUP: B53D2302625000).

\bibliographystyle{plainnat}
\bibliography{references}

\begin{thebibliography}{26}
\providecommand{\natexlab}[1]{#1}
\providecommand{\url}[1]{\texttt{#1}}
\expandafter\ifx\csname urlstyle\endcsname\relax
  \providecommand{\doi}[1]{doi: #1}\else
  \providecommand{\doi}{doi: \begingroup \urlstyle{rm}\Url}\fi

\bibitem[Campbell et~al.(2023)Campbell, Harvey, Weilbach, Bortoli, Rainforth, and Doucet]{jump_diffusion}
Andrew Campbell, William Harvey, Christian Weilbach, Valentin~De Bortoli, Tom Rainforth, and Arnaud Doucet.
\newblock Trans-dimensional generative modeling via jump diffusion models, 2023.
\newblock URL \url{https://arxiv.org/abs/2305.16261}.

\bibitem[Corso et~al.(2020)Corso, Cavalleri, Beaini, Li\`{o}, and Veli\v{c}kovi\'{c}]{corso2020principal}
Gabriele Corso, Luca Cavalleri, Dominique Beaini, Pietro Li\`{o}, and Petar Veli\v{c}kovi\'{c}.
\newblock Principal neighbourhood aggregation for graph nets.
\newblock In \emph{Advances in Neural Information Processing Systems}, volume~33, pages 13260--13271. Curran Associates, Inc., 2020.

\bibitem[Dhariwal and Nichol(2021)]{dhariwal2021diffusion}
Prafulla Dhariwal and Alexander~Quinn Nichol.
\newblock Diffusion models beat {GAN}s on image synthesis.
\newblock In A.~Beygelzimer, Y.~Dauphin, P.~Liang, and J.~Wortman Vaughan, editors, \emph{Advances in Neural Information Processing Systems}, 2021.
\newblock URL \url{https://openreview.net/forum?id=AAWuCvzaVt}.

\bibitem[Faez et~al.(2021)Faez, Ommi, Baghshah, and Rabiee]{faez2021survey}
Faezeh Faez, Yassaman Ommi, Mahdieh~Soleymani Baghshah, and Hamid~R. Rabiee.
\newblock Deep graph generators: A survey.
\newblock \emph{IEEE Access}, 9:\penalty0 106675--106702, 2021.
\newblock \doi{10.1109/ACCESS.2021.3098417}.

\bibitem[He et~al.(2021)He, You, Sandstr{\"{o}}m, Nittinger, Bjerrum, Tyrchan, Czechtizky, and Engkvist]{hysnjtce:1}
Jiazhen He, Huifang You, Emil Sandstr{\"{o}}m, Eva Nittinger, Esben~Jannik Bjerrum, Christian Tyrchan, Werngard Czechtizky, and Ola Engkvist.
\newblock Molecular optimization by capturing chemist's intuition using deep neural networks.
\newblock \emph{J. Cheminformatics}, 13\penalty0 (1):\penalty0 26, 2021.
\newblock \doi{10.1186/S13321-021-00497-0}.
\newblock URL \url{https://doi.org/10.1186/s13321-021-00497-0}.

\bibitem[Ho and Salimans(2022)]{hs:1}
Jonathan Ho and Tim Salimans.
\newblock Classifier-free diffusion guidance.
\newblock In \emph{NeurIPS 2021 Workshop DGMs Applications}, 2022.

\bibitem[Ho et~al.(2020)Ho, Jain, and Abbeel]{hja:1}
Jonathan Ho, Ajay Jain, and Pieter Abbeel.
\newblock Denoising diffusion probabilistic models.
\newblock In \emph{Advances in Neural Information Processing Systems}, volume~33, pages 6840--6851. Curran Associates, Inc., 2020.

\bibitem[Hoogeboom et~al.(2022)Hoogeboom, Satorras, Vignac, and Welling]{hoogeboom22a}
Emiel Hoogeboom, V\'{\i}ctor~Garcia Satorras, Cl{\'e}ment Vignac, and Max Welling.
\newblock Equivariant diffusion for molecule generation in 3{D}.
\newblock In \emph{Proceedings of the 39th International Conference on Machine Learning}, volume 162 of \emph{Proceedings of Machine Learning Research}, pages 8867--8887. PMLR, 17--23 Jul 2022.

\bibitem[Irwin and Shoichet(2005)]{is:1}
John~J. Irwin and Brian~K. Shoichet.
\newblock Zinc − a free database of commercially available compounds for virtual screening.
\newblock \emph{Journal of Chemical Information and Modeling}, 45\penalty0 (1):\penalty0 177--182, 2005.
\newblock \doi{10.1021/ci049714+}.
\newblock URL \url{https://doi.org/10.1021/ci049714}.
\newblock PMID: 15667143.

\bibitem[Jin et~al.(2018{\natexlab{a}})Jin, Barzilay, and Jaakkola]{jbj:1}
Wengong Jin, Regina Barzilay, and Tommi Jaakkola.
\newblock Junction tree variational autoencoder for molecular graph generation.
\newblock In \emph{Proceedings of the 35th International Conference on Machine Learning}, volume~80 of \emph{Proceedings of Machine Learning Research}, pages 2323--2332. PMLR, 10--15 Jul 2018{\natexlab{a}}.

\bibitem[Jin et~al.(2018{\natexlab{b}})Jin, Yang, Barzilay, et~al.]{jybj:1}
Wengong Jin, Kevin Yang, Regina Barzilay, et~al.
\newblock Learning multimodal graph-to-graph translation for molecular optimization.
\newblock In \emph{ICLR}, 2018{\natexlab{b}}.

\bibitem[Jin et~al.(2020)Jin, Barzilay, and Jaakkola]{jbj:2}
Wengong Jin, Regina Barzilay, and Tommi Jaakkola.
\newblock Hierarchical generation of molecular graphs using structural motifs.
\newblock In \emph{Proceedings of the 37th International Conference on Machine Learning}, ICML'20. JMLR.org, 2020.

\bibitem[Johnson et~al.(2021)Johnson, Austin, van~den Berg, and Tarlow]{johnson2021beyond}
Daniel~D. Johnson, Jacob Austin, Rianne van~den Berg, and Daniel Tarlow.
\newblock Beyond in-place corruption: Insertion and deletion in denoising probabilistic models.
\newblock In \emph{ICML Workshop on Invertible Neural Networks, Normalizing Flows, and Explicit Likelihood Models}, 2021.
\newblock URL \url{https://openreview.net/forum?id=cAsVBUe1Rnj}.

\bibitem[Ketata et~al.(2025)Ketata, Gao, Sommer, Wollschl{\"a}ger, and G{\"u}nnemann]{ketata2025lift}
Mohamed~Amine Ketata, Nicholas Gao, Johanna Sommer, Tom Wollschl{\"a}ger, and Stephan G{\"u}nnemann.
\newblock Lift your molecules: Molecular graph generation in latent euclidean space.
\newblock In \emph{The Thirteenth International Conference on Learning Representations}, 2025.

\bibitem[Kong et~al.(2023)Kong, Cui, Sun, Zhuang, Prakash, and Zhang]{DBLP:conf/icml/KongCSZPZ23}
Lingkai Kong, Jiaming Cui, Haotian Sun, Yuchen Zhuang, B.~Aditya Prakash, and Chao Zhang.
\newblock Autoregressive diffusion model for graph generation.
\newblock In Andreas Krause, Emma Brunskill, Kyunghyun Cho, Barbara Engelhardt, Sivan Sabato, and Jonathan Scarlett, editors, \emph{International Conference on Machine Learning, {ICML} 2023, 23-29 July 2023, Honolulu, Hawaii, {USA}}, volume 202 of \emph{Proceedings of Machine Learning Research}, pages 17391--17408. {PMLR}, 2023.
\newblock URL \url{https://proceedings.mlr.press/v202/kong23b.html}.

\bibitem[Liu et~al.(2018)Liu, Allamanis, Brockschmidt, and Gaunt]{Liu2018ConstrainedGV}
Qi~Liu, Miltiadis Allamanis, Marc Brockschmidt, and Alexander~L. Gaunt.
\newblock Constrained graph variational autoencoders for molecule design.
\newblock In \emph{Neural Information Processing Systems}, 2018.
\newblock URL \url{https://api.semanticscholar.org/CorpusID:43924638}.

\bibitem[Ninniri et~al.(2024)Ninniri, Podda, and Bacciu]{npb:1}
Matteo Ninniri, Marco Podda, and Davide Bacciu.
\newblock Classifier-free graph diffusion for molecular property targeting.
\newblock In Albert Bifet, Jesse Davis, Tomas Krilavi{\v{c}}ius, Meelis Kull, Eirini Ntoutsi, and Indr{\.{e}} {\v{Z}}liobait{\.{e}}, editors, \emph{Machine Learning and Knowledge Discovery in Databases. Research Track}, pages 318--335, Cham, 2024. Springer Nature Switzerland.
\newblock ISBN 978-3-031-70359-1.

\bibitem[Perez et~al.(2018)Perez, Strub, de~Vries, Dumoulin, and Courville]{psvdc:1}
Ethan Perez, Florian Strub, Harm de~Vries, Vincent Dumoulin, and Aaron Courville.
\newblock Film: Visual reasoning with a general conditioning layer.
\newblock \emph{Proceedings of the AAAI Conference on Artificial Intelligence}, 32\penalty0 (1), 2018.
\newblock \doi{10.1609/aaai.v32i1.11671}.

\bibitem[Ramakrishnan et~al.(2014)Ramakrishnan, Dral, Rupp, and von Lilienfeld]{Ramakrishnan2014}
Raghunathan Ramakrishnan, Pavlo~O. Dral, Matthias Rupp, and O.~Anatole von Lilienfeld.
\newblock Quantum chemistry structures and properties of 134 kilo molecules.
\newblock \emph{Scientific Data}, 1\penalty0 (1), August 2014.
\newblock ISSN 2052-4463.
\newblock \doi{10.1038/sdata.2014.22}.

\bibitem[Shabani et~al.(2023)Shabani, Hosseini, and Furukawa]{Shabani2023}
Mohammad~Amin Shabani, Sepidehsadat Hosseini, and Yasutaka Furukawa.
\newblock Housediffusion: Vector floorplan generation via a diffusion model with discrete and continuous denoising.
\newblock In \emph{2023 IEEE/CVF Conference on Computer Vision and Pattern Recognition (CVPR)}, page 5466–5475. IEEE, June 2023.
\newblock \doi{10.1109/cvpr52729.2023.00529}.
\newblock URL \url{http://dx.doi.org/10.1109/CVPR52729.2023.00529}.

\bibitem[Vignac et~al.(2023{\natexlab{a}})Vignac, Krawczuk, Siraudin, Wang, Cevher, and Frossard]{vkswcf:1}
Clement Vignac, Igor Krawczuk, Antoine Siraudin, Bohan Wang, Volkan Cevher, and Pascal Frossard.
\newblock Digress: Discrete denoising diffusion for graph generation.
\newblock In \emph{The Eleventh International Conference on Learning Representations}, 2023{\natexlab{a}}.

\bibitem[Vignac et~al.(2023{\natexlab{b}})Vignac, Osman, Toni, and Frossard]{midi}
Clement Vignac, Nagham Osman, Laura Toni, and Pascal Frossard.
\newblock Midi: Mixed graph and 3d denoising diffusion for molecule generation.
\newblock In \emph{Joint European Conference on Machine Learning and Knowledge Discovery in Databases}, pages 560--576. Springer, 2023{\natexlab{b}}.

\bibitem[You et~al.(2018{\natexlab{a}})You, Liu, Ying, Pande, and Leskovec]{you2018gpcn}
Jiaxuan You, Bowen Liu, Zhitao Ying, Vijay Pande, and Jure Leskovec.
\newblock Graph convolutional policy network for goal-directed molecular graph generation.
\newblock In \emph{Advances in Neural Information Processing Systems}, volume~31. Curran Associates, Inc., 2018{\natexlab{a}}.

\bibitem[You et~al.(2018{\natexlab{b}})You, Ying, Ren, et~al.]{you2018}
Jiaxuan You, Rex Ying, Xiang Ren, et~al.
\newblock {G}raph{RNN}: Generating realistic graphs with deep auto-regressive models.
\newblock In \emph{Proceedings of the 35th International Conference on Machine Learning}, volume~80 of \emph{Proceedings of Machine Learning Research}, pages 5708--5717. PMLR, 10--15 Jul 2018{\natexlab{b}}.

\bibitem[Zhao et~al.(2023)Zhao, Siriwardane, Wu, Fu, Al-Fahdi, Hu, and Hu]{Zhao2023}
Yong Zhao, Edirisuriya M.~Dilanga Siriwardane, Zhenyao Wu, Nihang Fu, Mohammed Al-Fahdi, Ming Hu, and Jianjun Hu.
\newblock Physics guided deep learning for generative design of crystal materials with symmetry constraints.
\newblock \emph{npj Computational Materials}, 9\penalty0 (1), March 2023.
\newblock ISSN 2057-3960.
\newblock \doi{10.1038/s41524-023-00987-9}.
\newblock URL \url{http://dx.doi.org/10.1038/s41524-023-00987-9}.

\bibitem[Zhu et~al.(2022)Zhu, Du, Wang, Xu, Zhang, Liu, and Wu]{zhu2022a}
Yanqiao Zhu, Yuanqi Du, Yinkai Wang, Yichen Xu, Jieyu Zhang, Qiang Liu, and Shu Wu.
\newblock A survey on deep graph generation: Methods and applications.
\newblock In \emph{The First Learning on Graphs Conference}, 2022.
\newblock URL \url{https://openreview.net/forum?id=Im8G9R1boQi}.

\end{thebibliography}
%%%%%%%%%%%%%%%%%%%%%%%%%%%%%%%%%%%%%%%%%%%%%%

\clearpage
\newpage

\section*{NeurIPS Paper Checklist}

\begin{enumerate}

\item {\bf Claims}
    \item[] Question: Do the main claims made in the abstract and introduction accurately reflect the paper's contributions and scope?
    \item[] Answer: \answerYes{} % Replace by \answerYes{}, \answerNo{}, or \answerNA{}.
    \item[] Justification: We have introduced a new model called \textsc{GrIDDD}, which can perform insert and delete operations. The Results section shows how it can match and even surpass the baseline used despite being trained on an arguably more difficult problem.

\item {\bf Limitations}
    \item[] Question: Does the paper discuss the limitations of the work performed by the authors?
    \item[] Answer: \answerYes{} % Replace by \answerYes{}, \answerNo{}, or \answerNA{}.
    \item[] Justification: We have discussed the limitations of our model in the Conclusions. We have been as transparent as possible.

\item {\bf Theory assumptions and proofs}
    \item[] Question: For each theoretical result, does the paper provide the full set of assumptions and a complete (and correct) proof?
    \item[] Answer: \answerNA{} % Replace by \answerYes{}, \answerNo{}, or \answerNA{}.
    \item[] Justification: Our paper does not include theoretical results.

    \item {\bf Experimental result reproducibility}
    \item[] Question: Does the paper fully disclose all the information needed to reproduce the main experimental results of the paper to the extent that it affects the main claims and/or conclusions of the paper (regardless of whether the code and data are provided or not)?
    \item[] Answer: \answerYes{} % Replace by \answerYes{}, \answerNo{}, or \answerNA{}.
    \item[] Justification: We have been as complete as possible. Our experiments employ the same architecture of previously published models with the same hyperparameters (albeit with minor modifications discussed in the paper). Regardless, we have discussed the architecture in detail in the supplemental material.

\item {\bf Open access to data and code}
    \item[] Question: Does the paper provide open access to the data and code, with sufficient instructions to faithfully reproduce the main experimental results, as described in supplemental material?
    \item[] Answer: \answerYes{} % Replace by \answerYes{}, \answerNo{}, or \answerNA{}.
    \item[] Justification: We include the source code, as well as how to set up a Conda environment to run it, in the supplemental material. The dataset is downloaded automatically by the source code. We will make the source code publicly available in the case of an acceptance.

\item {\bf Experimental setting/details}
    \item[] Question: Does the paper specify all the training and test details (e.g., data splits, hyperparameters, how they were chosen, type of optimizer, etc.) necessary to understand the results?
    \item[] Answer: \answerYes{} % Replace by \answerYes{}, \answerNo{}, or \answerNA{}.
    \item[] Justification: we have discussed the dataset splits in the supplemental material. As stated in the main paper, most of our hyperparameters are the same used by FreeGress, and we have only listed the differences with our solution. Our source code contains the various configurations files in .yaml format.

\item {\bf Experiment statistical significance}
    \item[] Question: Does the paper report error bars suitably and correctly defined or other appropriate information about the statistical significance of the experiments?
    \item[] Answer: \answerYes{} % Replace by \answerYes{}, \answerNo{}, or \answerNA{}.
    \item[] Justification: We reported the same metrics used by the baseline.

\item {\bf Experiments compute resources}
    \item[] Question: For each experiment, does the paper provide sufficient information on the computer resources (type of compute workers, memory, time of execution) needed to reproduce the experiments?
    \item[] Answer: \answerYes{} % Replace by \answerYes{}, \answerNo{}, or \answerNA{}.
    \item[] Justification: We discussed the type of GPU employed. In the appendix, we discussed the run time compared to the baseline. The runtime is similar for all experiments, so we did not find it necessary to report every single run time.
    
\item {\bf Code of ethics}
    \item[] Question: Does the research conducted in the paper conform, in every respect, with the NeurIPS Code of Ethics \url{https://neurips.cc/public/EthicsGuidelines}?
    \item[] Answer: \answerYes{} % Replace by \answerYes{}, \answerNo{}, or \answerNA{}.
    \item[] Justification: We believe our work is conforming to the Code of Ethics.

\item {\bf Broader impacts}
    \item[] Question: Does the paper discuss both potential positive societal impacts and negative societal impacts of the work performed?
    \item[] Answer: \answerNA{} % Replace by \answerYes{}, \answerNo{}, or \answerNA{}.
    \item[] Justification: We do not believe that our work will have any societal impact.
    
\item {\bf Safeguards}
    \item[] Question: Does the paper describe safeguards that have been put in place for responsible release of data or models that have a high risk for misuse (e.g., pretrained language models, image generators, or scraped datasets)?
    \item[] Answer: \answerNA{} % Replace by \answerYes{}, \answerNo{}, or \answerNA{}.
    \item[] Justification: Our work poses no such risks.

\item {\bf Licenses for existing assets}
    \item[] Question: Are the creators or original owners of assets (e.g., code, data, models), used in the paper, properly credited and are the license and terms of use explicitly mentioned and properly respected?
    \item[] Answer: \answerYes{} % Replace by \answerYes{}, \answerNo{}, or \answerNA{}.
    \item[] Justification: We have credited the paper on which we based our own code. We have cited both ZINC-250k and QM9. We are not aware of any license or terms of use concerning these assets.

\item {\bf New assets}
    \item[] Question: Are new assets introduced in the paper well documented and is the documentation provided alongside the assets?
    \item[] Answer: \answerNA{} % Replace by \answerYes{}, \answerNo{}, or \answerNA{}.
    \item[] Justification: We did not release any new asset.

\item {\bf Crowdsourcing and research with human subjects}
    \item[] Question: For crowdsourcing experiments and research with human subjects, does the paper include the full text of instructions given to participants and screenshots, if applicable, as well as details about compensation (if any)? 
    \item[] Answer: \answerNA{} % Replace by \answerYes{}, \answerNo{}, or \answerNA{}.
    \item[] Justification: Our work does not involve crowdsourcing and we did not perform research with human subjects.

\item {\bf Institutional review board (IRB) approvals or equivalent for research with human subjects}
    \item[] Question: Does the paper describe potential risks incurred by study participants, whether such risks were disclosed to the subjects, and whether Institutional Review Board (IRB) approvals (or an equivalent approval/review based on the requirements of your country or institution) were obtained?
    \item[] Answer: \answerNA{} % Replace by \answerYes{}, \answerNo{}, or \answerNA{}.
    \item[] Justification: Our work does not involve crowdsourcing and we did not perform research with human subjects.

\item {\bf Declaration of LLM usage}
    \item[] Question: Does the paper describe the usage of LLMs if it is an important, original, or non-standard component of the core methods in this research? Note that if the LLM is used only for writing, editing, or formatting purposes and does not impact the core methodology, scientific rigorousness, or originality of the research, declaration is not required.
    %this research? 
    \item[] Answer: \answerNA{} % Replace by \answerYes{}, \answerNo{}, or \answerNA{}.
    \item[] Justification: The core method development in this research does not involve LLMs as any important, original, or non-standard components

\end{enumerate}

\newpage
\appendix

%\newpage

\section{Additional details on the methodology} 
\subsection{Delete scheduler and insert/delete timestep distributions} \label{app:zeta}
\begin{figure}[!htb]
    \begin{minipage}{0.49\textwidth}
        \includegraphics[width=1\textwidth]{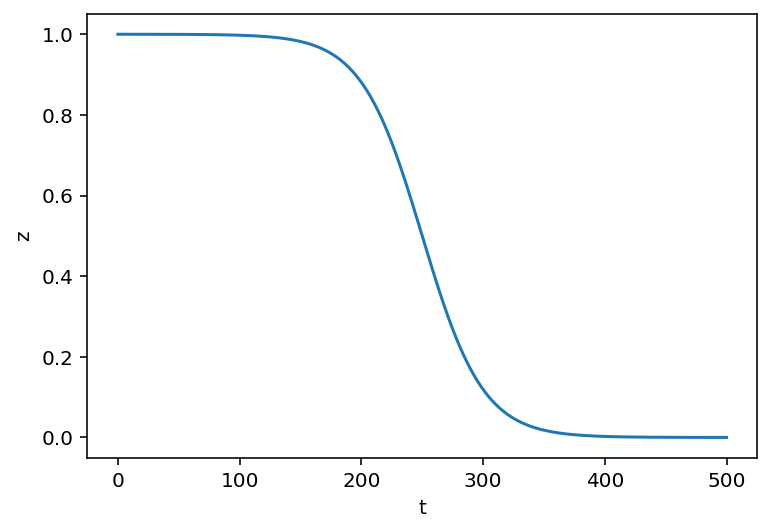}
        \subcaption{\label{fig:zeta} $\zeta(t)$}
    \end{minipage}
    \begin{minipage}{0.49\textwidth}
        \includegraphics[width=1\textwidth]{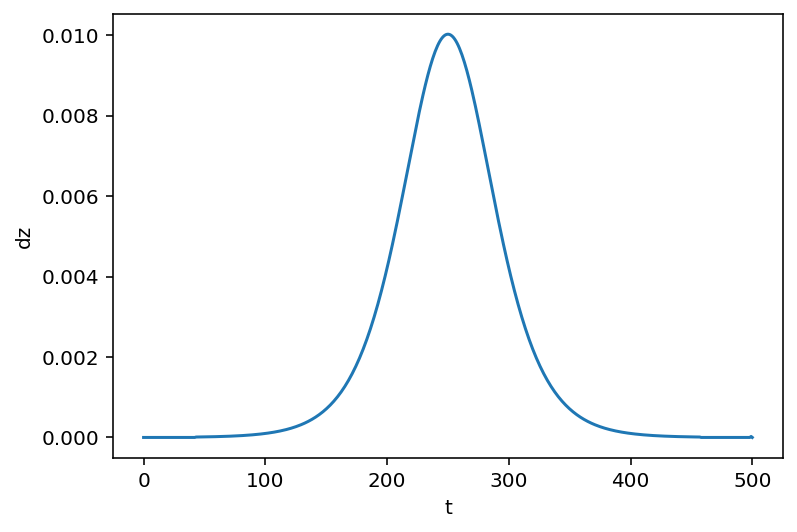}
        \subcaption{\label{fig:d_zeta} $\zeta^{'}(t) = \frac{\delta\zeta(t)}{\delta t}$}
    \end{minipage}
    
    \caption{\label{fig:zeta_t} The functions $\zeta(t)$ and its derivative $\zeta^{'}(t)$ for $w = 0.05$ and $T = 500$. $1- \zeta(t)$ is effectively the ``delete scheduler'', and represents the probability that a node selected for deletion has already been deleted at timestep $t$. $\zeta^{'}(t)$ represents instead the probability that a node chosen for deletion will switch to category \texttt{DEL}$^{*}$ exactly at timestep $t$.}
\end{figure}

We show a possible realization of the delete scheduler $\zeta(t)$ and the insert/delete timestep distribution $\zeta^{'}(t)$ in Figure \ref{fig:zeta_t}.

\subsection{Distribution of the labels of the inserted nodes and edges} \label{app:labels}
During the forward process, \textsc{GrIDDD} assigns labels to the newly inserted nodes and edges using a marginal distribution of these labels computed on the individual training sample. Such distributions can be computed just once before starting the training process and can also be saved for multiple training sessions. 
We have also tried different distributions during our preliminary studies, such as always inserting the least frequent node/edge class, as well as sampling them from a uniform probability, but the sample’s marginal distribution gave us better results.

\subsection{Training and sampling algorithms} \label{app:algos}
Here, we provide the pseudocode describing the procedures to train (Algorithm \ref{alg:new_training}) and sample from (Algorithm \ref{alg:new_denoising}) \textsc{GrIDDD}.

\begin{algorithm}
\caption{Pseudocode of the training algorithm} \label{alg:new_training}
\begin{algorithmic}[1] 

\Require $\mathcal{G} = \{\bm{G}_i\}_{i=1}^{N}$, dataset of $N$ graphs
\Require $T$, the number of denoising timesteps,
\Require $p_\theta$, the main model,
\Require $g_\phi$, the model tasked to predict the number of \texttt{DEL}$^{*}$s in $\bm{G}^t$,
\Require $\zeta^{'}(t)$, a discrete probability distribution function defined on $T$,
\Require $h(n)$, which given a graph size $n$ returns a probability distribution $h_{n}$ on the number of nodes.

\ForEach {graph $\bm{G}_i = (\bm{X}_i, \bm{E}_i) \in \mathcal{G}$:}
    \State $n^0 = |\bm{X}_i|$
    \State $n^T \sim h_{n^0}(n)$
    \State $t \sim Uniform(1,T)$
    \State $\bm{m}_{\bm{X}_i}, \bm{m}_{\bm{E}_i}$ = marginal distribution computed on the node and edges of $\bm{G}_i$.
    \State $\bm{S}^{*} = \{0, \dots, 0\}$ vector storing the insertion times of the nodes
    \State $\bm{X}_i^{*t}, \bm{E}_i^{*t} \sim (\bm{X}_i^t)^{'}\overline{Q}_{\bm{X}}^{t|0},  (\bm{E}^t_i)^{'}\overline{Q}_{\bm{E}}^{t|0}$ 
    \State $\Delta^T = n^T - n^0$ 
    \State Sample $\bm{U} = \{u_j\}$ timesteps, with $j\in\{0 \dots |\Delta^T|-1\}$, where $u_j \sim \zeta^{'}$
    \State Remove from $\bm{U}$ the elements greater than $t$
    \ForEach{$u_j \in \bm{U}$} 
        \If{$\Delta^T > 0$} \Comment{Insert}
            \State Sample a new node $\bm{x} \sim \bm{m}_{\bm{X}_i}$ with insertion time $u_j$, and add it to $\bm{X}_i$
            \State Sample $\bm{x}^t$ from the distribution $\bm{x}^{'}\overline{\bm{Q}}^{*t|u_j}_{\bm{X}}$ and add it to $\bm{X}_i^{*t}$
            \State Sample from $\bm{m}_{\bm{E}_i}$, for each node already in the graph, an inward and 
            \Statex outward edge with the newly added $\bm{x}_t$ 
            \State Sample the edges $\bm{e}$'s final state $\bm{e}^t$ from $\bm{e}^{'}\overline{Q}^{*t|u_j}_{\bm{E}}$ and add them to $\bm{E}_i^{*t}$
            \State Append $u_j$ at the end of $\bm{S}^{*}$ 
        \ElsIf{$\Delta^T < 0$} \Comment{Delete}
            \State Sample one random node $\bm{x}$ among the ones in $\bm{G}_i$ that are not set to state \texttt{DEL} 
            \Statex or \texttt{DEL}$^{*}$
            \If{$u_j = t$}
                \State $\bm{x} = $ \texttt{DEL}$^{*}$
                \State Set all incoming and outgoing edges of $\bm{x}$ as \texttt{DEL}$^{*}$
            \Else
                \State Remove $\bm{x}$ from $\bm{X}_i^{*t}$
                \State Remove all incoming and outgoing edges of $\bm{x}$ from $\bm{E}_i^{*t}$
                \State Remove the entry in $\bm{S}^{*}$ associated to the node
            \EndIf
        \EndIf
    \EndFor
    
    \State $\bm{G}^{*t} = (\bm{X}^{*t}, \bm{E}^{*t})$
    \State $\{(\widehat{\bm{X}}^{*0}, \widehat{\bm{E}}^{*0}), \widehat{\bm{S}}^{*}\} = p_\theta(\bm{G}^{*t})$
    \State $loss_i = \lambda_X \mathrm{CE}(\widehat{\bm{X}}^{*0}, \bm{X}^{*0}) + \lambda_E \mathrm{CE}(\widehat{\bm{E}}^{*0}, \bm{E}^{*0}) + \lambda_S \mathrm{CE}(\widehat{\bm{S}}^{*0}, \bm{S}^{*0})$ 
    \State $n_{\texttt{DEL}^{*}} = $ number of nodes set to \texttt{DEL}$^{*}$ in $\bm{G}^{*t}$
    \State Remove all nodes set to \texttt{DEL}$^{*}$ in $\bm{G}^{*t}$ as well as the associated edges
    \State $loss^{*}_i = CE(g_\phi(\bm{G}^{*t}), n_{\texttt{DEL}^{*}})$ 
\EndFor
\end{algorithmic}
\end{algorithm}

\begin{algorithm}
\caption{Pseudocode of the denoising algorithm} \label{alg:new_denoising}
\begin{algorithmic}[1] 

\Require $T$, the number of denoising timesteps,
\Require $n^T$, the number of nodes in the noisy graph $\bm{G}^T$,
\Require $p_\theta$, the main model,
\Require $g_\phi$, the model tasked to predict the number of \texttt{DEL}$^{*}$s in $\bm{G}^t$,
\Require $\bm{m_X}$ and $\bm{m_E}$, the dataset's marginal distributions on the nodes and the edges;

\State Sample $\bm{G}^T = (\bm{X}^T, \bm{E}^T)$ with $n^T$ nodes from an the marginal distributions $\bm{m_X}$ and $\bm{m_E}$;

\ForEach {timestep $t \in [T, \ldots, 1]$:}
    \State Use $g_{\phi}(\bm{G}^t)$ to predict the number of \texttt{DEL}$^{*}$ entries to insert in the graph, and insert them.
    \State Set all the edges connecting the newly inserted nodes to the rest of the graph as \texttt{DEL}$^{*}$.
    \State Use $p_\theta(\bm{G}^t)$ to predict $\bm{S}^t$, the insertion times for each node in the graph, and $\bm{G}^{*0}$.
    \State Remove the nodes $\bm{x}_i$ such that $\bm{S}_i = t$, as well as their edges.
	\State Use $\bm{S}^t$ and $\bm{G}^{*0}$ to sample, for each node $\bm{x}_i^{t}$ and edge $\bm{e}_{ij}^{t}$ in $\bm{G}^{t}$, their type at step $t-1$:
    \State $\bm{x}_i^{t-1} \sim p_\theta(\bm{x}_i^{t-1}|\bm{x}_i^{t}) 
= \sum_{x \in \chi} \frac{q(\bm{x}_i^{t}|\bm{x}_i^{t-1}) q(\bm{x}_i^{t-1}|\bm{x}_i^{{S}_i} = x)}{q(\bm{x}_i^{t}|\bm{x}_i^{{S}_i} = x)}p_\theta(\bm{x}_i^{{S}_i} = x| \bm{x}^t)$.
    \State $\bm{e}_{ij}^{t-1} \sim p_\theta(\bm{e}_{ij}^{t-1}|\bm{e}_{ij}^{t}) 
= \sum_{\bm{e} \in \varepsilon} \frac{q(\bm{e}_{ij}^{t}|\bm{e}_{ij}^{t-1}) q(\bm{e}_{ij}^{t-1}|\bm{e}_{ij}^{max({S}_i, {S}_j)} = \bm{e})}{q(\bm{e}_{ij}^{t}|\bm{e}_{ij}^{max(\bm{S}_i, \bm{S}_j)} = \bm{e})}p_\theta(\bm{e}_{ij}^{max({S}_i, {S}_j)} = \bm{e}| \bm{e}_{ij}^t)$.
    \State $\bm{G}^{t-1} = (\{ \bm{x}_i^{t-1} \}, \{ \bm{e}_{ij}^{t-1} \})$
\EndFor
\end{algorithmic}
\end{algorithm}

\section{Additional experimental details}

\subsection{Dataset splits}\label{app:splits}
\paragraph{QM9.} On QM9, the training set is made of the first 100000 samples in the dataset. The test set is made of 10$\%$ of the overall data, and the remaining data is used to make the validation set.

\paragraph{ZINC-250k. } The training set uses the first 80$\%$ of the data. The remainder is equally split between the validation set and the test set.

\subsection{Neural architecture} \label{sec:architecture}
\begin{figure*}[!ht]

\tikzset{every picture/.style={line width=0.75pt}}%set default line width to 0.75pt      
\centering

\begin{subfigure}{.36\textwidth}
\centering
\resizebox{!}{6.5cm}{
\includegraphics{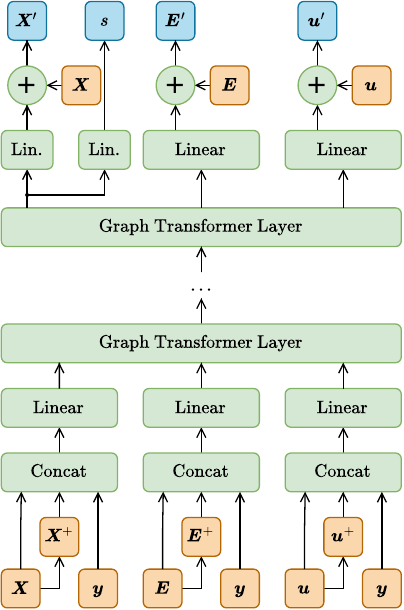}}
\caption{}\label{fig:architecture}
\end{subfigure}
\begin{subfigure}{.165\textwidth}
\centering
\resizebox{!}{6.5cm}{
\includegraphics{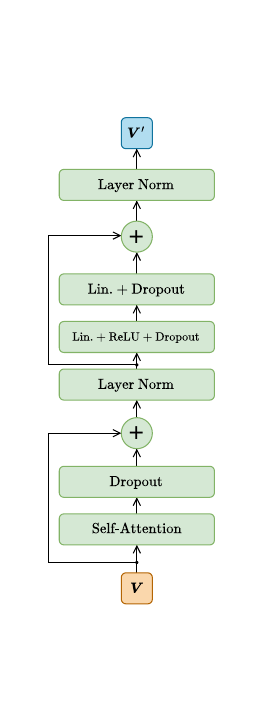}}
\caption{}\label{fig:graph-transformer}
\end{subfigure}
\begin{subfigure}{.46\textwidth}
\centering
\resizebox{!}{6.5cm}{
\includegraphics{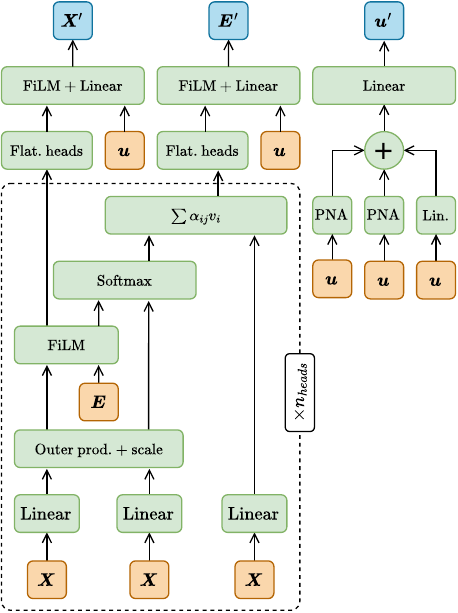}}
\caption{}\label{fig:self-attention}
\end{subfigure}
\caption{(a) High-level architecture of \textsc{GrIDDD}. (b) Detail of the graph transformer layer. $\bm{V}$ represents the triple $(\bm{X}, \bm{E}, \bm{u})$. (c) The self attention-layer within the graph transformer layer. In the pictures, orange boxes indicate inputs, blue boxes indicate outputs, and green boxes indicate layers/operations.}
\label{fig:digress_nn}
\end{figure*}

Both the main model and the auxiliary model employed to predict the number of \texttt{DEL}s are based on FreeGress' architecture. The model is composed of a stack of Graph Transformer layers. It receives as input the node features matrix $\bm{X}$, edge features matrix $\bm{{E}}$, and a vector $\bm{u}$ encoding the timestep $t$ and optional extra features. These inputs are augmented with conditioning information $\bm{y}$ (denoted with the $+$ superscript in Figure~\ref{fig:architecture}) and processed through the transformer stack.

Each Graph Transformer layer (Figure~\ref{fig:graph-transformer}) follows the standard architecture: self-attention, dropout, residual connection, layer normalization, and a feedforward block. The self-attention mechanism (Figure~\ref{fig:self-attention}) applies scaled dot-product attention to the augmented node features $\bm{X}$. The resulting attention weights are modulated by the edge tensor $\bm{{E}}$ via a FiLM layer \citep{psvdc:1}, normalized with softmax, and used to re-weight $\bm{X}$. The output is flattened and combined with $\bm{u}$ through another FiLM layer before linear projection to prediction logits.

Separately, $\bm{X}$ and $\bm{{E}}$ are passed through independent PNA layers \citep{corso2020principal}, and their outputs are summed with a linearly projected $\bm{u}$ to form the final graph representation. In \textsc{GrIDDD}, the final output $\bm{u}^{'}$ is employed by our auxiliary model to predict the amount of \texttt{DEL}$^{*}$ nodes that must be reintroduced at the current timestep. The final Graph Transformer Layer's output designed to predict the matrix $\bm{X}^{'}$ is passed to two Linear layers, instead of just one. The first is used, just like before, to predict the final node matrix. The second one is used instead to predict the activation timestep $s$ of the various nodes.

\subsection{Baselines} \label{app:baseline}
For the task of property targeting, discussed in Section \ref{sec:targeting}, we compared our model against DiGress \citep{vkswcf:1}, a Graph Diffusion Model capable of performing conditional graph generation through a classifier-based guidance, and FreeGress \citep{npb:1}, a Graph Diffusion Model which employs a classifier-free guidance system instead.
For the task of property optimization, discussed in Section \ref{sec:optimization}, we employ a baseline of three models. In particular, we employ the Junction-Tree Variational Autoencoder (JT-VAE, \cite{jbj:1}), an Autoencoder that works on the molecular sub-structural level rather than the atomic level, the Graph Convolutional Policy Network (GCPN, \cite{you2018gpcn}), which employs Reinforcement Learning to generate molecules atom-by-atom, and the Constrained Graph Variational Autoencoders (CG-VAE, \cite{Liu2018ConstrainedGV}), a graph-based Autoencoder whose peculiarity is that it applies chemical validity constraints during the generative process to increase its performance in molecular tasks.

\subsection{Considerations on the independence between nodes and edges in the forward process} \label{app:node_edge_indip}
When a node is switched to $\text{\texttt{DEL}}^{*}$, so are the edges associated with it. This effectively creates a dependence between the two, while standard Diffusion Models assume a mutual independence between them. This is similar to what is done with Masked Diffusion \citep{DBLP:conf/icml/KongCSZPZ23}, where the edges associated with masked nodes are masked as well. In our case, what we effectively do is replacing the forward process $p(\bm{e}_{ij}^{t}|\bm{e}_{ij}^{t-1})$ with $p(\bm{e}_{ij}^{t}|\bm{e}_{ij}^{t-1}, \bm{x}_{i}^{t},  \bm{x}_{j}^{t})$, and $p(\bm{e}_{ij}^{t-1}|\bm{e}_{ij}^{t}, \bm{e}_{ij}^{0})$ with $p(\bm{e}_{ij}^{t-1}|\bm{e}_{ij}^{t}, \bm{e}_{ij}^{0}, \bm{x}_{i}^{t}, \bm{x}_{j}^{t})$. However, it should be kept in mind how the latter equation is, in all practical scenarios, identical to $p(\bm{e}_{ij}^{t-1}|\bm{e}_{ij}^{t}, \bm{e}_{ij}^{0})$, as the only situations where $\bm{x}_{i/j}^{t}$ could influence the value of $\bm{e}_{ij}^{t-1}$ is when the former is labeled as $\text{\texttt{DEL}}^{*}$ and the latter to $\text{\texttt{DEL}}$, in which case $\bm{e}_{ij}^{t-1}$ should switch to $\text{\texttt{DEL}}^{*}$ as well. However, this scenario is not possible in practice, as neither the nodes nor the edges can appear explicitly with label $\text{\texttt{DEL}}$ during the reverse process. We leave to future research the study of independent edge processes that still allow for the deletion of the associated nodes.

\section{Additional computational considerations}

\subsection{Runtime}\label{app:runtime}
On average, one training epoch of \textsc{GrIDDD} takes 309.1 seconds on two nVidia A100 GPUs. FreeGress, on the same problem (hence, without insert and delete operations), takes 243 seconds. This means that \textsc{GrIDDD} is approximately 27$\%$ slower than its counterpart. This is similar to the 33.3$\%$ increase we have found on QM9, where \textsc{GrIDDD} takes 18 seconds on average to complete a training epoch on a single A100, while FreeGress takes 13.5.

\subsection{Sampling speed}\label{app:sampling-speed}
Interestingly enough, \textsc{GrIDDD} can be faster at sampling than standard Diffusion Models. When tasked with generating a variegated batch of graphs with different sizes, a standard Diffusion Model would first sample the graph sizes from the dataset's node distribution. This means that, in practice, the sizes of the node and edge matrices will have one or more dimensions equal to the highest graph size sampled. In turn, this means that every graph generated will require a processing power proportional to such size.

On the other hand, \textsc{GrIDDD} is capable of generating a graph of any shape, even when starting from a different size. In this scenario, one could start the sampling process with very small latents (even zero nodes). Since the tensors representing the sample batch would be very small, the computational power required to process these tensors will be small as well. This means that the early phases of the sampling process, before most nodes are inserted, will likely be faster than standard diffusion (although this is highly dependent on $\zeta(t)$).

To test this hypothesis, we have sampled 128 molecules on ZINC-250K, conditioning on the LogP value, with FreeGress and \textsc{GrIDDD}.  For the latter, we have performed three experiments. In the first, \textsc{GrIDDD} starts from latents with a size of two nodes. In the second, we have used an initial size of 24 (the most frequent graph size in the training set), and then we have sampled the initial graph size using the training set's node distribution, as is done in FreeGress. The results, summarized in Table \ref{tab:sampling_speed}, show how \textsc{GrIDDD} is capable of performing as well as FreeGress while halving the sampling time in the best case scenario, and even in the worst case scenario it requires as little as five percent more time to perform as well as FreeGress.

\begin{table}[!htbp]
    \centering
    %\begin{tabular}{l|cccccccc}
    \caption{Sampling speed comparison between GrIDDD and FreeGress when generating 128 samples on ZINC-250k conditioned on the LogP value, using different initial latent sizes.}
    \begin{tabular}{lccc}
    \toprule
         Model & \textbf{Validity $\uparrow$} & \textbf{MAE $\downarrow$} & \textbf{Sampling time $\downarrow$}\\ 
         \midrule 
         FreeGress                         & 89.8$\%$ & 0.31 & 152.05s\\ 
         \textsc{GrIDDD} (initial size=2)           & $\bm{91.4\%}$ & $\bm{0.23}$ &  $\bm{68.09s}$\\ 
         \textsc{GrIDDD} (initial size=24)          & $90.6\%$ & 0.31 & 120.19s\\
         \textsc{GrIDDD} (initial size $\sim p(n)$) & $88.2\%$ & 0.26 & 159.85s\\
        \bottomrule
    \end{tabular}
    \label{tab:sampling_speed}
\end{table}

\section{Additional results}
\subsection{Comparison with EDM-SyCO} \label{app:syco}
We compared to the 3D Diffusion Model by \cite{ketata2025lift} called EDM-SyCO, since it performs property optimization similarly to \textsc{GrIDDD}. Notably, EDM-SyCO also inputs 3D coordinates to the diffusion process.
To ensure a fair comparison, we slightly adapted our original setup to the less restrictive one used in the related paper. In particular, they attempt to optimize each molecule 100 times, rather than 20 as in our previous experiment. Then, among the ones within the similarity threshold, they select the 10 ones with the best improvement, duplicate each one of them ten times, and run the optimization process a second time. The process is repeated four times (for a total of 400 optimization rounds), after which the molecule with the best improvement within the similarity threshold is selected. All the successful molecules obtained in the process (that is, within the similarity threshold) are considered when computing the diversity score. In cases where the number of successful experiments is less than two, the diversity score is set to zero. Results of the comparison are reported in Table \ref{tab:optimization-syco}, showing competitive or superior performances by \textsc{GrIDDD}, even though it does not benefit from using 3D information as input. We have compared our results against EDM-SyCo on property targeting as well. Specifically, we have trained a model on ZINC-250K  with 1000 denoising timesteps (as is done by our baseline) that is conditioned on the molecular weight. We have obtained a MAE of $2.13 \pm 0.19$ and a validity of $86.2\%$, while EDM-SyCo records a MAE of $3.86 \pm 0.08$ and a validity of $88\%$.

\begin{table}[ht]
\caption{Comparison with EDM-SyCo on property optimization.}
\centering
\footnotesize
\renewcommand{\arraystretch}{1.2}
\begin{tabular}{@{}lcccccccc@{}}
\toprule
\multirow{2}{*}{\textbf{Method}} & 
\multicolumn{2}{c}{\textbf{LogP (sim $\geq$ 0.4)}} & 
\multicolumn{2}{c}{\textbf{LogP (sim $\geq$ 0.6)}} & 
\multicolumn{2}{c}{\textbf{QED (sim $\geq$ 0.4)}} & 
\multicolumn{2}{c}{\textbf{DRD2 (sim $\geq$ 0.4)}} \\

\cmidrule{2-9}
& \textbf{Improv. $\uparrow$} & \textbf{Div.$\uparrow$} & \textbf{Improv. $\uparrow$} & \textbf{Div.$\uparrow$} & \textbf{Succ.$\uparrow$} & \textbf{Div.$\uparrow$} & \textbf{Succ.$\uparrow$} & \textbf{Div.$\uparrow$} \\
\midrule
\textsc{EDM-SyCo}    & 3.11 $\pm$ 1.27 & \textbf{0.555}  & 1.51 $\pm$ 1.10 & \textbf{0.360} & 46.4 $\%$ & 0.163 & \textbf{27.3 $\%$} & \textbf{0.083} \\

\midrule
\textsc{GrIDDD} &  \textbf{3.27 $\pm$ 0.91} & 0.511 &  \textbf{1.59 $\pm$ 0.54} & \textbf{0.359} & \textbf{64.1 $\%$}  & \textbf{0.269} & 19.7 $\%$ & 0.058 \\
\bottomrule
\end{tabular}
\label{tab:optimization-syco}
\end{table}

\subsection{Hyperparameter sensitivity analysis}\label{app:sensitivity-analysis}
We show in Table \ref{tab:hyperparams} an analysis of \textsc{GrIDDD}'s sensitivity of the hyperparameters regulating the scheduler ($D$ and $w$) and the node counts distribution ($p_{\text{min}}$ and $p_{\text{max}}$). One interesting insight that can be gathered from the table is the fact that the number of split molecules generated significantly increases with smaller $D$s. From a practical perspective, small values for this hyperparameter imply that, during the denoising process, the model can insert nodes relatively late during the reverse process. We conjecture that split molecules are mostly caused by nodes inserted too late during such a process, as they cannot be re-attached in time to the main graph since the noise scheduler does not allow for too many changes in the graph’s structure during the last phases of the process. We have also noted that this phenomenon is sensibly reduced when \textsc{GrIDDD} is given in input, as extra features, different powers of the adjacency matrix.
\begin{table}[!htbp]
    \centering
    \footnotesize
    \setlength{\tabcolsep}{5.25pt}
    \caption{Study of GrIDDD's sensitivity to hyperparameter changes when sampling 100 molecules after training on the QM9 dataset. The hyperparameters tested are the delete scheduler's parameters $D, w$, and the node count distribution's parameters $p_{\text{min}}$ and $p_{\text{max}}$. Val: validity, Avg NC: average number of connected components, Max NC: maximum number of connected components, NSC: number of graphs sampled with a single connected component.}
    \parbox{.49\linewidth}{
    \begin{tabular}{lcccc}
        \toprule
        $D$ & \textbf{Val $\uparrow$} & \textbf{Avg NC $\downarrow$} & \textbf{Max NC $\downarrow$} & \textbf{NSC $\uparrow$}\\ 
        \midrule 
        0.25 & 0.93 & 1.15 & 4 & 88 \\ 
        0.50 & 0.93 & 1.02 & 2 & 98 \\ 
        0.75 & 0.95 & 1.01 & 2 & 99 \\
        \midrule
        $p_{\text{min}}$ & \textbf{Val $\uparrow$} & \textbf{Avg NC $\downarrow$} & \textbf{Max NC $\downarrow$} & \textbf{NSC $\uparrow$} \\ 
        \midrule 
        0.2 & 0.93 & 1.02 & 2 & 98  \\ 
        0.4 & 0.91 &  1   & 1 & 100 \\ 
        0.6 & 0.96 & 1.02 & 3 & 99  \\
        0.8 & 0.94 & 1.02 & 2 & 98  \\
        \bottomrule
    \end{tabular}
    }
    \setlength{\tabcolsep}{5.25pt}
    \parbox{.49\linewidth}{
    \begin{tabular}{lcccc}
		\toprule
        $w$ & \textbf{Val $\uparrow$} & \textbf{Avg NC $\downarrow$} & \textbf{Max NC $\downarrow$} & \textbf{NSC $\uparrow$}\\ 
        \midrule 
        0.025 & 0.94 &  1   & 1 & 100 \\ 
        0.050 & 0.93 & 1.02 & 2 & 98  \\ 
        0.075 & 0.92 & 1.03 & 2 & 97  \\
        \midrule
        $p_{\text{max}}$ & \textbf{Val $\uparrow$} & \textbf{Avg NC $\downarrow$} & \textbf{Max NC $\downarrow$} & \textbf{NSC $\uparrow$} \\ 
        \midrule 
        0.5   & 0.86 & 1.02 & 2 & 98 \\ 
        0.1   & 0.93 & 1.02 & 2 & 98 \\ 
        0.01  & 0.89 & 1.01 & 2 & 99 \\
        0.001 & 0.94 & 1.02 & 2 & 98 \\
        \bottomrule
    \end{tabular}
    }
    \label{tab:hyperparams}
\end{table}

\subsection{Generated molecules}\label{app:samples}
\begin{figure}[ht]
    \centering
    \includegraphics[width=1\linewidth]{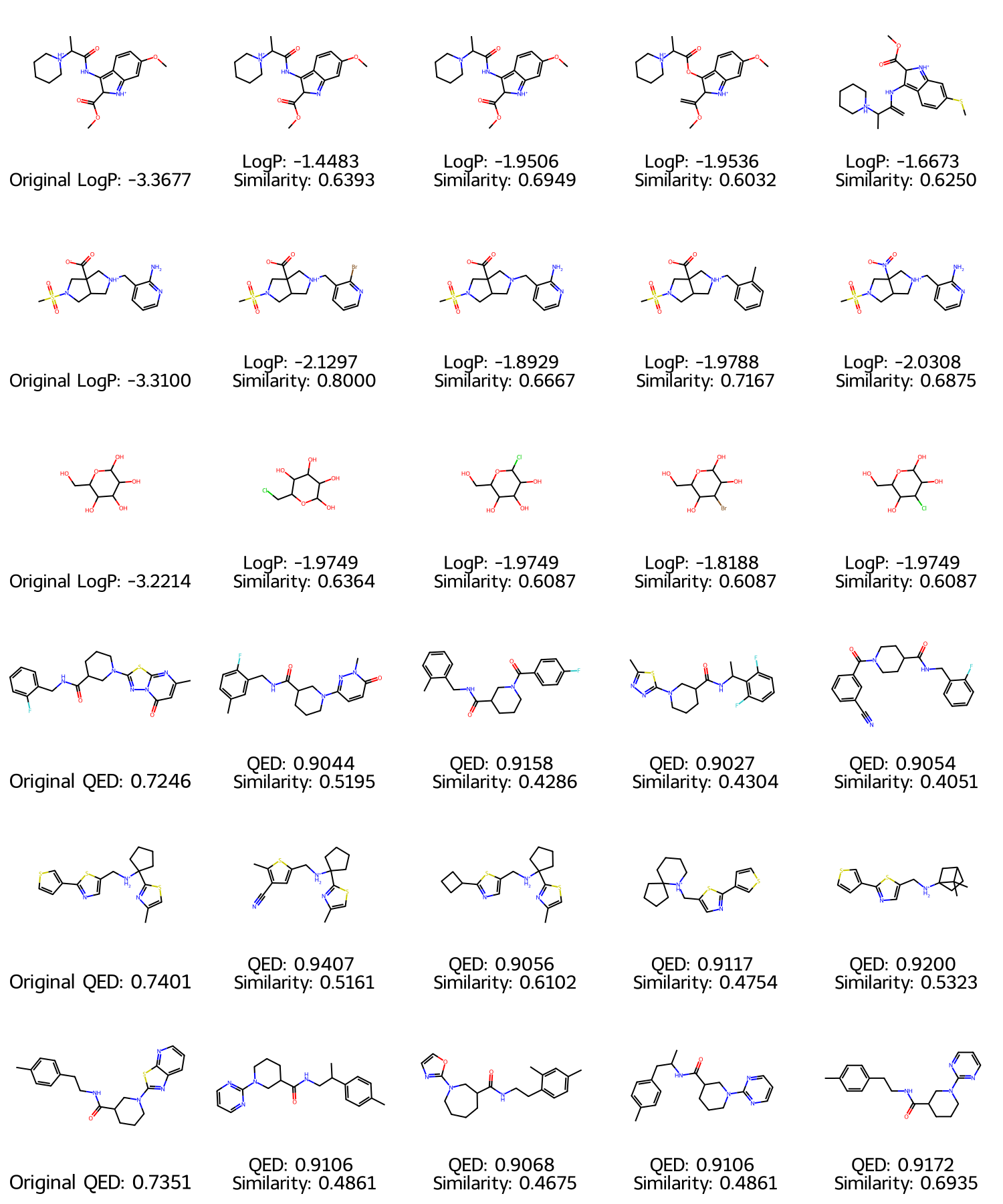}
    \caption{Randomly selected optimization experiments on ZINC-250k. The first element in each row is the original molecule, while the other four are results of different optimization experiments. We report, for each experiment, the original value and the resulting one.}
    \label{fig:zinc_optimization}
\end{figure}

\begin{figure}[ht]
    \centering
    \includegraphics[width=1\linewidth]{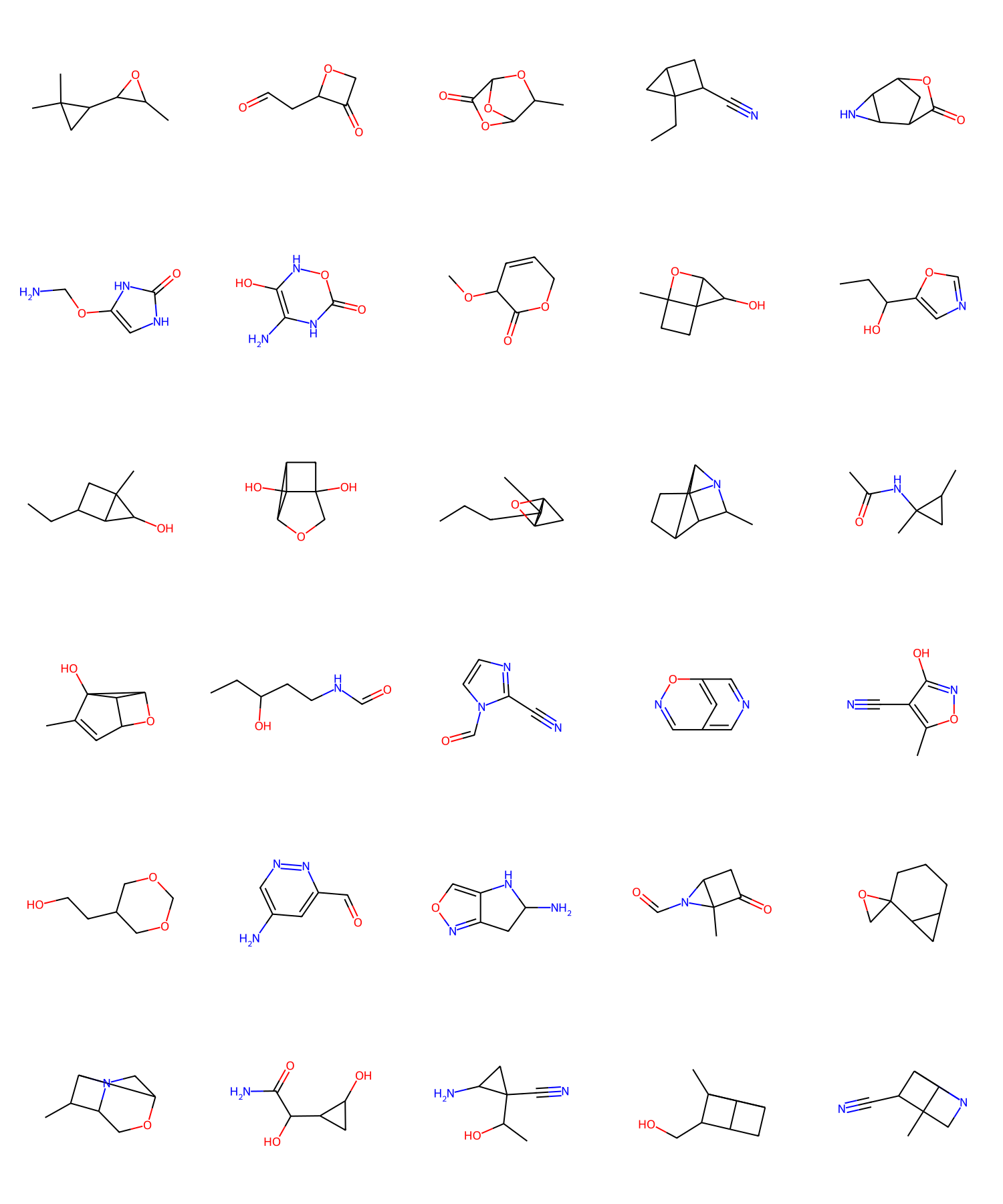}
    \caption{30 non-curated samples generated without conditioning on QM9.}
    \label{fig:qm9_mols}
\end{figure}

Figure \ref{fig:zinc_optimization} shows 6 (3 for the QED task, 3 for the LogP task) randomly selected molecules optimized by \textsc{GrIDDD} on the ZINC-250k dataset. For each molecule, we show 4 different successful optimizations. Figure \ref{fig:qm9_mols} shows 30 random molecules generated without conditioning by \textsc{GrIDDD} on QM9.
%\newpage

\subsection{Failure Analysis} \label{app:failure}

While analyzing the situations where \textsc{GrIDDD} fails in conditional generation the most, we have noticed how, unlike FreeGress, the experiments targeting $\mu$ tend to fail most frequently when targeting values close to zero. The few molecules generated tend to be small, split graphs. We have investigated the phenomenon and found out that the molecule with SMILES string \texttt{CC} (ethane) is known to have a dipole moment equal to zero. We believe that it is likely that \textsc{GrIDDD} tries to generate ethane or similar small molecules to minimize $\mu$, with higher failure rates than usual since these small molecules are under-represented in the dataset. Remarkably, DiGress and FreeGress are unable to pursue this minimization, since they almost inevitably start the reverse process from latents with a larger graph size and cannot adapt it to produce smaller molecules later on. Overall, this is an interesting behavior which shows that \textsc{GrIDDD} can successfully learn the properties of molecules with infrequent sizes as well.

%%%%%%%%%%%%%%%%%%%%%%%%%%%%%%%%%%%%%%%%%%%%%%%%%%%%%%%%%%%%

\end{document}